%% file: main.tex
\input{_constants}
\arxiv 

\pdfoutput=1
\documentclass[10pt,twocolumn,letterpaper]{article}
\input{cvpr_header}

\unless\ifarxiv \myexternaldocument{_supplementary} \fi

\begin{document}
\title{Fourier-Guided Attention Upsampling for Image Super-Resolution}
\author{
    Daejune Choi    \qquad
    Youchan No      \qquad
    Jinhyung Lee    \qquad
    Duksu Kim
    \\
    Korea University of Technology and Education (KOREATECH) \\
    {\tt\small \{eowns13, bluebear102, refracta, bluekds\}@koreatech.ac.kr} \\
    {\small \url{https://hpc-lab-koreatech.github.io/FGA-SR/}}
}
\maketitle

\input{sec/00_abstract}
\input{sec/01_intro}
\input{sec/02_related}

\input{sec/03_method}
\input{sec/04_experiment}

\input{sec/10_conclusion}
{\small
\bibliographystyle{ieeenat_fullname}
\bibliography{sec/11_references}
}

\ifarxiv \clearpage \appendix \input{sec/12_appendix} \fi

\end{document}

%% file: _constants.tex
\def\paperID{0000}
\def\confName{ICCV}
\def\confYear{2025}

\newif\ifreview 
\newif\ifarxiv \newcommand{\arxiv}{\arxivtrue}
\newif\ifcamera 
\newif\ifrebuttal 

%% file: cvpr_header.tex
\ifreview \usepackage[review]{cvpr} \fi
\ifarxiv \usepackage[pagenumbers]{cvpr} \fi
\ifrebuttal \usepackage[rebuttal]{cvpr} \fi
\ifcamera \usepackage{cvpr} \fi

\input{_macros}  

\usepackage{xr-hyper}

\makeatletter
\newcommand*{\addFileDependency}[1]{
  \typeout{(#1)}
  \@addtofilelist{#1}
  \IfFileExists{#1}{}{\typeout{No file #1.}}
}

\makeatother
\newcommand*{\myexternaldocument}[1]{
    \externaldocument{#1}
    \addFileDependency{#1.tex}
    \addFileDependency{#1.aux}
}

\definecolor{cvprblue}{rgb}{0.21,0.49,0.74}
\usepackage[pagebackref,breaklinks,colorlinks,allcolors=cvprblue]{hyperref}
\usepackage[capitalize]{cleveref}
\crefname{section}{Sec.}{Secs.}
\crefname{table}{Table}{Tables}
\crefname{figure}{Fig.}{Figs.}

\ifarxiv \crefname{appendix}{App.}{Apps.}
\else \crefname{appendix}{Suppl.}{Suppls.} \fi

%% file: _macros.tex
\usepackage{graphicx}	
\usepackage{amsmath}	
\usepackage{amssymb}	
\usepackage{booktabs}
\usepackage{times}
\usepackage{microtype}
\usepackage{epsfig}
\usepackage{caption}
\usepackage{float}
\usepackage{placeins}
\usepackage{color, colortbl}
\usepackage{stfloats}
\usepackage{enumitem}
\usepackage{tabularx}
\usepackage{xstring}
\usepackage{multirow}
\usepackage{xspace}
\usepackage{url}
\usepackage{subcaption}
\usepackage{xcolor}
\usepackage[hang,flushmargin]{footmisc}

\usepackage{makecell}
\usepackage{adjustbox}
\usepackage{cuted}
\ifcamera \usepackage[accsupp]{axessibility} \fi

% ...

\ifarxiv  \fi

\newcommand{\R}[1]{{%
    \textbf{%
        \ifstrequal{#1}{1}{\textcolor{red}{R#1}}{%
        \ifstrequal{#1}{2}{\textcolor{blue}{R#1}}{%
        \ifstrequal{#1}{3}{\textcolor{magenta}{R#1}}{%
        \ifstrequal{#1}{4}{\textcolor{teal}{R#1}}{%
                           \textcolor{cyan}{R#1}%
        }}}}%
    }%
}}

%% file: sec/00_abstract.tex
\begin{abstract}

We propose Fourier-Guided Attention (FGA), a lightweight upsampling module for single image super-resolution. 
Conventional upsamplers, such as Sub-Pixel Convolution, are efficient but frequently fail to reconstruct high-frequency details and introduce aliasing artifacts.
FGA addresses these issues by integrating (1) a Fourier feature-based Multi-Layer Perceptron (MLP) for positional frequency encoding, (2) a cross-resolution Correlation Attention Layer for adaptive spatial alignment, and (3) a frequency-domain L1 loss for spectral fidelity supervision.
Adding merely 0.3M parameters, FGA consistently enhances performance across five diverse super-resolution backbones in both lightweight and full-capacity scenarios.
Experimental results demonstrate average PSNR gains of 0.12–0.14 dB and improved frequency-domain consistency by up to 29\%, particularly evident on texture-rich datasets.
Visual and spectral evaluations confirm FGA's effectiveness in reducing aliasing and preserving fine details, establishing it as a practical, scalable alternative to traditional upsampling methods.
\end{abstract}

%% file: sec/01_intro.tex
\section{Introduction}
\label{sec:intro}
\input{figs/feat_results_edsr}
Single image super-resolution (SISR) aims to reconstruct a high-resolution (HR) image from a low-resolution (LR) input, recovering fine textures, sharp edges, and structural details that are typically degraded during downsampling.
Driven by deep learning, recent advances in convolutional neural networks (CNNs)~\cite{SRCNN, Conv-FSRCNN, Conv-SRDenseNet, Conv-VDSR, Conv-EDSR, Conv-RDN, ConvAttn-RCAN, Conv-RFANet, ConvAttn-SAN, ConvAttn-RNAN, ConvAttn-HAN, ConvAttn-NLSN} and Transformer-based architectures~\cite{IPT, SwinIR, CAT, EDT, Uformer, DAT, HAT, DRCT} have significantly pushed the frontier of SISR, yielding impressive performance on widely used benchmarks.

Despite these advances, restoring high-frequency details such as repetitive textures, edges, and fine contours remains challenging, as such features are often severely degraded or completely absent in LR images.
The quality of SISR outcomes is critically dependent on the ability of the upsampling module to reconstruct plausible high-frequency information.
Traditional approaches like interpolation-based methods, transposed convolution (deconvolution), and the widely used Sub-Pixel Convolution (SPC)~\cite{ESPCN} commonly introduce spatial artifacts, notably checkerboard patterns and spectral aliasing, due to their inherent spatial and frequency insensitivity (Sec.~\ref{sec:analysis_artifacts} and Fig.~\ref{fig:feat_results_edsr}).

To address these limitations, we propose the Fourier-Guided Attention (FGA) upsampling module—a lightweight, frequency-aware solution specifically designed to enhance high-frequency fidelity and reduce reconstruction artifacts  (Sec.~\ref{sec:method}).
FGA integrates: (1) a Fourier Feature-based Multi-Layer Perceptron (FF-MLP) that explicitly encodes positional and spectral information, (2) a Correlation Attention Layer (CAL) that adaptively aligns HR features with LR contexts through cross-resolution attention, and (3) a frequency-domain L1 loss (FL1) that supervises spectral fidelity directly.
Our extensive experiments demonstrate that FGA consistently improves both quantitative metrics and perceptual quality across various state-of-the-art SISR backbones and diverse benchmarks, making it a versatile and effective replacement for conventional upsampling techniques (Sec.~\ref{sec:experiment}).

%% file: figs/feat_results_edsr.tex
\begin{figure*}[!t]
    \renewcommand{\arraystretch}{1.1}
    \renewcommand{\tabcolsep}{0.9mm}
    \centering
    \resizebox{\textwidth}{!}{%
        \begin{tabular}{ccccccccccc}
            \phantomsubcaption\label{fig:feat_spec_edsr}
            & &  Interp+Conv &  Deconv &  PixelShuffle &  Ours (FGA) &&&  GT &  HR (bottom) / LR (top) \\
            \multirow{2}{*}[15pt]{\rotatebox{90}{\textbf{(a) Spectrum}}}
            & \multirow{1}{*}[38pt]{\rotatebox{90}{ Pre-}}
            & \includegraphics[height=0.15\linewidth, width=0.15\linewidth]{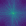}
            & \includegraphics[height=0.15\linewidth, width=0.15\linewidth]{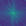}
            & \includegraphics[height=0.15\linewidth, width=0.15\linewidth]{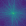}
            & \includegraphics[height=0.15\linewidth, width=0.15\linewidth]{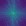}
            & \multicolumn{1}{c|}{}
            & \multicolumn{1}{|c}{}
            & \includegraphics[height=0.15\linewidth, width=0.15\linewidth]{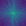} 
            & \multirow{2}{*}[65pt]{\centering\includegraphics[height=0.31\linewidth, keepaspectratio=true]{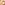}} \\
            
            & \multirow{1}{*}[40pt]{\rotatebox{90}{ Post-}}
            & \includegraphics[height=0.15\linewidth, width=0.15\linewidth]{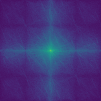} 
            & \includegraphics[height=0.15\linewidth, width=0.15\linewidth]{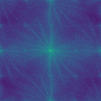}
            & \includegraphics[height=0.15\linewidth, width=0.15\linewidth]{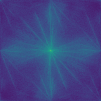}
            & \includegraphics[height=0.15\linewidth, width=0.15\linewidth]{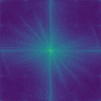}
            & \multicolumn{1}{c|}{} 
            & \multicolumn{1}{|c}{}
            & \includegraphics[height=0.15\linewidth, width=0.15\linewidth]{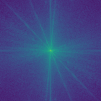} 
            & \\  [15pt]

            \phantomsubcaption\label{fig:feat_map_edsr}
            \multirow{2}{*}[20pt]{\rotatebox{90}{\textbf{(b) Feature map}}} 
            & \multirow{1}{*}[38pt]{\rotatebox{90}{ Pre-}}
            & \includegraphics[height=0.15\linewidth, width=0.15\linewidth]{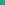}
            & \includegraphics[height=0.15\linewidth, width=0.15\linewidth]{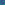}
            & \includegraphics[height=0.15\linewidth, width=0.15\linewidth]{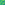}
            & \includegraphics[height=0.15\linewidth, width=0.15\linewidth]{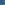}
            & \multicolumn{1}{c|}{} 
            & \multicolumn{1}{|c}{}
            & \includegraphics[height=0.15\linewidth, width=0.15\linewidth]{imgs/features/EDSR/feat/cropped_img_007_lr.pdf} 
            & \multirow{2}{*}[65pt]{\centering\includegraphics[height=0.31\linewidth, keepaspectratio=true]{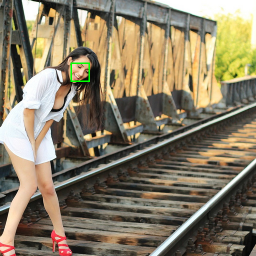}} \\
            
            & \multirow{1}{*}[40pt]{\rotatebox{90}{ Post-}}
            & \includegraphics[height=0.15\linewidth, width=0.15\linewidth]{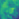}
            & \includegraphics[height=0.15\linewidth, width=0.15\linewidth]{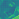}
            & \includegraphics[height=0.15\linewidth, width=0.15\linewidth]{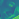}
            & \includegraphics[height=0.15\linewidth, width=0.15\linewidth]{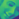}
            & \multicolumn{1}{c|}{} 
            & \multicolumn{1}{|c}{}
            & \includegraphics[height=0.15\linewidth, width=0.15\linewidth]{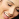} 
            & & 
        \end{tabular}
    }
\caption{%
Feature map and frequency spectrum analysis of different upsampling methods using the EDSR backbone on ``img\_007'' from Urban100~\cite{Urban100}.
The input is center-cropped to $128 \times 128$ pixels and upsampled by a factor of $\times4$.
The top two rows visualize feature maps before and after upsampling; the bottom two rows show their corresponding Fourier spectra.
}
\label{fig:feat_results_edsr}
\end{figure*}

%% file: sec/02_related.tex
\section{Related Work}\label{sec:related}

\subsection{SISR with Deep Neural Networks}\label{subsec:sisr-dnn}

Deep neural networks have significantly advanced the field of SISR, learning complex mappings from LR to HR images and surpassing traditional interpolation methods.
These advancements can be broadly categorized by network architectures, specifically CNN-based methods, attention mechanisms, and Transformer-based models.

Early CNN-based methods, such as SRCNN~\cite{SRCNN}, established the effectiveness of deep learning in SISR by outperforming classical interpolation.
Subsequent approaches improved upon this foundation by increasing network depth and incorporating advanced convolutional designs~\cite{ResNet, SRGAN, ESRGAN, Conv-RDN}.
Notably, EDSR~\cite{Conv-EDSR} refined residual learning by removing redundant components like batch normalization layers, significantly enhancing efficiency and performance.
Meanwhile, RDN~\cite{Conv-RDN} introduced Residual Dense Blocks (RDBs), leveraging densely connected convolutional layers and a contiguous memory mechanism for rich local feature extraction and fusion.
However, CNN-based models still face challenges in restoring critical high-frequency details necessary for sharp, detailed image reconstruction.

Attention mechanisms have addressed some of these limitations by enabling models to prioritize salient features, capture long-range dependencies, and effectively reconstruct HR images.
Attention-enhanced models can more comprehensively capture global contexts compared to purely CNN-based counterparts~\cite{Attention, ConvAttn-SAN, ConvAttn-HAN, ConvAttn-RNAN, ConvAttn-RCAN, ConvAttn-NLSN, RecursiveSR-nonlocal}.
For example, RCAN~\cite{ConvAttn-RCAN} utilizes a deep residual-in-residual (RIR) architecture paired with channel attention, adaptively recalibrating channel features and emphasizing informative regions for superior discriminative learning.

Transformer architectures, originally introduced for natural language processing, have recently shown promising results for SISR~\cite{SwinIR, IPT, EDT, HAT, Uformer, DAT, CAT, DRCT}.
IPT~\cite{IPT} employs a Transformer-based encoder-decoder framework with multi-task pre-training, achieving excellent performance across various low-level vision tasks.
SwinIR~\cite{SwinIR}, building on the Swin Transformer~\cite{SwinTransformer}, adopts self-attention within fixed windows and shifted window strategies to enlarge receptive fields, surpassing many CNN-based methods.
Models like Uformer~\cite{Uformer}, CAT~\cite{CAT}, HAT~\cite{HAT}, and DRCT~\cite{DRCT} further innovate Transformer architectures through diverse window attentions, feature aggregation, and dense connections to achieve state-of-the-art image restoration results.

While recent Transformer-based SISR models primarily enhance encoder structures to capture complex representations, our proposed method innovatively focuses on the decoder.
Specifically, we enhance the widely adopted pixel-shuffle method by integrating frequency-domain features into the decoding phase, significantly improving reconstruction fidelity and high-frequency detail recovery.

\subsection{Frequency-Domain Guided SISR Methods}
Recent advances in SISR have leveraged frequency-domain information to enhance reconstruction quality, particularly by explicitly incorporating global frequency cues through techniques such as wavelet transforms and discrete cosine transforms (DCT).
For instance, DWSR~\cite{DWSR} applies wavelet transforms to decompose images into low- and high-frequency sub-bands and utilizes CNNs to combine these sub-bands effectively for reconstruction. 
WaveletSRNet~\cite{WaveletSRNet} introduces a wavelet-based loss function for predicting high-resolution coefficients from low-resolution images, specifically tailored to face image super-resolution.
MWCNN~\cite{MWCNN} integrates multi-level wavelet transforms within a U-Net architecture, enhancing multiscale representation capabilities for various image restoration tasks.
In addition, Fourier Space Loss introduced by Fuoli et al.~\cite{fuoli2021fourier} directly supervises high-frequency content restoration in the Fourier domain, significantly improving perceptual image quality. 
WaveMixSR\cite{WaveMixSR} proposes a frequency-aware token-mixing strategy based on wavelet decomposition, achieving state-of-the-art results on the BSD100×2 benchmark~\cite{BSDS100}.
Furthermore, ORDSR~\cite{ORDSR} integrates DCT operations within convolutional layers, facilitating smooth spatial-to-frequency domain transitions.

Unlike prior works, which predominantly embed frequency information in encoder modules, our approach distinctly incorporates frequency-domain features directly into the decoder.
Our novel design exploits Fourier features explicitly during the decoding phase, offering enhanced reconstruction fidelity and superior high-frequency detail recovery.

\subsection{Upsampling Modules in SISR Networks}

Upsampling modules play a critical role in SISR, enhancing the spatial resolution of feature maps.
The choice of upsampling technique significantly affects both the quality and computational efficiency of SISR models.
Generally, upsampling methods can be categorized into interpolation-based techniques, deconvolution layers, and pixel-shuffle methods.

Interpolation methods, such as bilinear or bicubic interpolation, represent the simplest approach, generating missing pixels based on neighboring pixel intensities.
Dong et al.~\cite{SRCNN} employed interpolation followed by convolutional layers to refine details.
Similarly, Odena et al.~\cite{NNConv} combined nearest-neighbor interpolation with convolutional layers, achieving smoother reconstructions with fewer visual artifacts.
Deconvolution layers, or transposed convolutions~\cite{Deconv}, offer a learnable upsampling solution by adapting filters from training data.
However, these methods often suffer from checkerboard artifacts, necessitating meticulous architectural designs and additional techniques to mitigate such issues.

Pixel-shuffle, introduced by ESPCN~\cite{ESPCN}, rearranges channels of feature maps into expanded spatial dimensions, making it computationally efficient and widely adopted across recent SISR models~\cite{Conv-EDSR, Conv-RDN, SRGAN, ESRGAN, ConvAttn-SAN, ConvAttn-HAN, ConvAttn-RCAN, ConvAttn-RNAN, ConvAttn-NLSN, SwinIR, IPT, EDT, CAT, DAT, HAT, DRCT}.
Additionally, recent arbitrary-scale super-resolution techniques have introduced innovative upsampling modules~\cite{INR_0, INR_1, INR_2, MetaSR, ArbRCAN, LIIF, LTE}, yet pixel-shuffle remains prevalent for fixed-scale super-resolution.

In this work, we focus on introducing a novel upsampling module specifically designed for fixed-scale SR networks, aiming to overcome the limitations of existing approaches and further enhance reconstruction quality.

%% file: sec/03_method.tex
\section{Analysis of Artifacts in Existing Upsampling Modules} \label{sec:analysis_artifacts}

Early SISR models commonly employed pre-upsampling techniques, which performed feature extraction in the high-resolution domain~\cite{SRCNN, Conv-VDSR, RecursiveSR-DRCN, DRFN}.
However, these methods incurred substantial computational overhead due to operating on large spatial resolutions. 
Consequently, post-upsampling strategies gained popularity, highlighting the critical role of efficient and effective upsampling modules in modern SISR pipelines~\cite{Conv-EDSR, Conv-RDN, SRGAN, ESRGAN, ConvAttn-SAN, ConvAttn-HAN, ConvAttn-RCAN, ConvAttn-RNAN, ConvAttn-NLSN, SwinIR, IPT, EDT, CAT, DAT, HAT, DRCT}.

Upsampling modules in SISR can be classified into traditional interpolation-based methods, and learned approaches such as deconvolution (transposed convolution) and sub-pixel convolution (PixelShuffle).
Interpolation methods, like bicubic interpolation, estimate new pixel values from neighboring pixels based on predefined kernels.
Although computationally simple and efficient, these methods inherently lack the ability to restore high-frequency details, resulting in smooth and blurred outputs~\cite{Park2003, RecursiveSR-DRCN, ESPCN, LapSRN}.

Deconvolution layers, which leverage learnable kernels, adaptively perform upsampling but often introduce checkerboard artifacts due to uneven kernel overlap, causing spectral aliasing and undesirable visual grid patterns~\cite{NNConv}.
Such artifacts significantly undermine image reconstruction fidelity.

PixelShuffle (PS)~\cite{ESPCN}, or sub-pixel convolution, rearranges channels of low-resolution feature maps into spatial dimensions, creating high-resolution outputs.
Due to its computational efficiency and seamless integration into end-to-end SISR frameworks, PS has become widely used, such as EDSR~\cite{Conv-EDSR}.
Nevertheless, PS can produce checkerboard artifacts closely tied to spectral aliasing~\cite{LCTC}.

Fig.\ref{fig:feat_results_edsr} compares upsampling methods—interpolation-based convolution (Interp+Conv), PS, and deconvolution (Deconv)—within the EDSR backbone, maintaining identical network components and training configurations for fairness.
Models were trained on the DIV2K dataset for 500k iterations, with evaluations conducted on the Urban100 test set\cite{Urban100}.

Fig.~\ref{fig:feat_spec_edsr} visualizes the frequency spectra obtained via Fast Fourier Transform (FFT), comparing encoder outputs (pre-upsampling) and upsampled feature maps (post-upsampling).
The spectra of traditional methods show evident artifacts, notably replicated frequency patterns indicative of high-frequency component folding into basebands, highlighting the issue of spectral aliasing.
These frequency-domain artifacts often manifest spatially as checkerboard patterns.
Fig.~\ref{fig:feat_map_edsr} further illustrates these artifacts through PCA-based dimensionality reduction visualizations of encoder and upsampled feature maps at ×4 scale.

The observations underscore limitations of existing upsampling methods, motivating our proposal for a robust, frequency-aware alternative.
As depicted in the \textit{Ours (FGA)} column of Fig.~\ref{fig:feat_results_edsr}, our approach effectively mitigates spectral aliasing and checkerboard artifacts, achieving cleaner spatial feature representations and more coherent frequency distributions.

\section{Fourier-Guided Attention Upsampling}\label{sec:method}

In this section, we first provide an overview of the architecture of our proposed upsampler.
We then describe the key components and design principles that form the foundation of our method.

\subsection{Overview of the FGA upsampler}
\input{figs/fga_architecture}

We propose a Fourier-Guided Attention (FGA) upsampler that enhances high-resolution reconstruction by leveraging frequency-aware representations and attention-based spatial refinement.
As illustrated in Fig.~\ref{fig:fga_architecture}, the FGA module is positioned after the backbone network and operates directly on its output feature maps.
Unlike conventional upsampling methods, FGA is explicitly designed to mitigate spectral aliasing and preserve fine-grained textures through frequency-domain encoding and adaptive correlation modeling.

The FGA upsampler consists of two main components: (1) a Fourier Feature-based Multi-Layer Perceptron (FF-MLP), and (2) a Correlation Attention Layer (CAL).
The core structure includes $N$ repeated stages, each comprising a convolutional layer, FF-MLP, and PixelShuffle unit, followed by a CAL module. A final convolutional layer is applied to produce the HR output.

The FF-MLP module enhances feature representations by injecting Fourier positional encodings derived from spatial coordinates $(x, y)$.
These encoded features are combined with the intermediate features from the previous stage and processed through an MLP to improve spatial discrimination and sensitivity to high-frequency content.

The CAL module performs attention-based refinement by computing query-key correlations between the upsampled features and backbone features.
This allows the network to adaptively emphasize informative regions and correct spatial inconsistencies introduced during upsampling.

Together, these components enable FGA to effectively suppress aliasing artifacts and enhance the fidelity of reconstructed textures.
The entire module is fully differentiable and end-to-end trainable, making it compatible with a wide range of SISR backbone architectures.

\subsection{Fourier Feature-based MLP}\label{subsec:ffmlp}

The pixel-shuffle upsampler, though efficient, inherently struggles to reconstruct high-frequency details. 
Because the convolutional layers preceding pixel-shuffle are unaware of the precise spatial destinations of the sub-pixels they generate, feature channels are often interleaved without positional distinction. 
This lack of spatial conditioning causes identical high-frequency components to be replicated across sub-pixels, producing artifacts such as repeated textures and aliasing.

To address these limitations, we enhance the pixel-shuffle block with a shallow MLP guided by explicit positional information. 
Inspired by Fourier-feature positional encodings from implicit neural representations~\cite{INR_0, INR_1, INR_2} and their adaptation to arbitrary-scale SR~\cite{LTE}, we inject normalized 2D coordinate embeddings into the MLP. 
Unlike arbitrary-scale methods that embed HR coordinates during inference, our design is tailored for fixed-scale SR by pre-aligning each sub-pixel feature with its designated spatial position in the output grid.

For efficiency, we align the coordinate encoding process with the pixel-shuffle pipeline. 
Rather than computing HR-level embeddings for every pixel, we partition the feature tensor into distinct sub-pixel groups and assign each group a unique coordinate identity in the low-resolution space. 
This ensures positional conditioning without incurring redundant computations.

\paragraph{Coordinate-Aware Feature Embedding}
Let $F \in \mathbb{R}^{\tfrac{H}{r} \times \tfrac{W}{r} \times C}$ denote the backbone feature map before upsampling, where $r$ is the upsampling factor.
We begin by generating a normalized HR coordinate grid $v \in \mathbb{R}^{H \times W \times 2}$, and apply \texttt{PixelUnshuffle} to align these coordinates with the sub-pixel layout:
\begin{equation}
V = \operatorname{PixelUnshuffle}(v), 
\quad V \in \mathbb{R}^{\tfrac{H}{r} \times \tfrac{W}{r} \times r^2 \times 2}
\end{equation}

Each of the $r^2$ sub-pixel groups is assigned its own spatial identity vector, which we encode via Fourier features.
These coordinate encodings are fused with the backbone features through element-wise multiplication to produce a frequency-aware representation:
\begin{equation}
F_{\text{ff}} = F \odot 
\begin{bmatrix}
\cos(\pi \cdot F V) \\
\sin(\pi \cdot F V)
\end{bmatrix}
\end{equation}

The modulated feature $F_{\text{ff}}$ is then passed through an MLP:
\begin{equation}
F_{\text{MLP}} = \operatorname{MLP}(F_{\text{ff}})
\end{equation}

Finally, the pixel-shuffle operation reconstructs the upsampled feature map:
\begin{equation}
F_{\uparrow} = \operatorname{PixelShuffle}(F_{\text{MLP}}),
\quad F_{\uparrow} \in \mathbb{R}^{H \times W \times C}
\end{equation}

This structure ensures that each sub-pixel channel is modulated based on its specific spatial destination, enabling the MLP to learn distinct high-frequency behaviors for different spatial locations.
As a result, our approach effectively suppresses aliasing and reduces checkerboard artifacts.

\paragraph{Relation to Transformer-Style Multi-Head}
Our use of sub-pixel spatial heads bears conceptual similarity to multi-head attention in transformers. Whereas transformer heads learn independent channel-wise relationships, our spatial heads learn position-dependent bases via Fourier encoding.
Each head processes only the coordinate features corresponding to a specific sub-pixel group, allowing the MLP to generate spatially diverse outputs and minimizing high-frequency replication across sub-pixels.

\subsection{Correlation Attention Layer}
\label{subsec:CAL}
To further mitigate aliasing artifacts and correct spatial inconsistencies introduced by naive upsampling, we introduce the Correlation Attention Layer (CAL).
Unlike typical upsampling approaches that rely solely on learned features in HR space, CAL explicitly reuses the original LR context by enabling the HR features to selectively attend to their corresponding LR features through localized attention.
This module complements the frequency-aware enhancement of FF-MLP by focusing on spatial alignment and content refinement.

\paragraph{Window-Based Cross-Resolution Attention}
Let $F_{\mathrm{LR}} \in \mathbb{R}^{\tfrac{H}{r} \times \tfrac{W}{r} \times C}$ and $F_{\mathrm{HR}} \in \mathbb{R}^{H \times W \times C}$ denote the feature maps before and after upsampling by a scale factor $r$, respectively.
Following the window-based attention strategy from SwinIR~\cite{SwinIR}, we partition both features into non-overlapping windows of size $(\tfrac{M}{r}, \tfrac{M}{r})$ for the LR domain and $(M, M)$ for the HR domain.
This results in $N = \tfrac{HW}{M^2}$ corresponding window pairs, denoted as
\[
F_{\mathrm{LR}}^w \in \mathbb{R}^{N \times \tfrac{M^2}{r^2} \times C}, \quad
F_{\mathrm{HR}}^w \in \mathbb{R}^{N \times M^2 \times C}.
\]

Within each window, we perform cross-resolution attention: the HR feature window $w_\uparrow$ acts as the query, and the corresponding LR window $w$ provides the keys and values.
The attention mechanism and subsequent refinement are defined as:
\begin{equation}
\begin{aligned}
Q &= w_\uparrow, \quad K = w P_K, \quad V = w P_V, \\
w_\uparrow &= \operatorname{XRA}(w, w_\uparrow) + w_\uparrow, \\
w_\uparrow &= \operatorname{MLP}(w_\uparrow) + w_\uparrow,
\end{aligned}
\label{eq:attention}
\end{equation}
where $P_K$ and $P_V$ are learnable linear projection matrices, and $\operatorname{XRA}$ denotes the cross-resolution attention operation.
This mechanism enables the HR representation to incorporate structural guidance from its corresponding LR window, thereby improving spatial consistency and mitigating upsampling-induced artifacts.

Compared to conventional self-attention in the HR domain, CAL offers a more efficient solution by exploiting the reduced spatial size of LR features for key-value computation.
The computational complexity of standard self-attention (SA) is:
\begin{equation}
\Omega(\mathrm{SA}) = 4HWC^2 + 2M^2HWC,
\end{equation}
while the complexity of CAL becomes:
\begin{equation}
\Omega(\mathrm{CA}) = \left(1 + \frac{2}{r^2} \right) HWC^2 + \frac{2M^2}{r^2} HWC.
\end{equation}

\paragraph{Overlapping Window Attention}
While the Swin Transformer~\cite{SwinIR} mitigates the limited receptive field of local attention by stacking multiple shifted-window blocks, such stacking leads to a quadratic increase in computation with higher resolutions—making it unsuitable for lightweight SISR upsamplers.

To expand the receptive field more efficiently, we adopt the Overlapping Window Correlation Attention (OWCA) mechanism~\cite{HAT}, which enlarges spatial coverage using a single attention layer.
Specifically, we apply overlapping partitioning to the LR feature map before upsampling, allowing each LR window to access a wider context while still using fewer pixels than its HR counterpart.

For an overlap ratio $\alpha$, the window size for the LR domain becomes Eq.~\ref{eq:M0}.
\begin{equation}
M_O = (1 + \alpha) \cdot \frac{M}{r}
\label{eq:M0}
\end{equation}
%
While overlapping increases the receptive field, it results in only a moderate computational overhead:
\begin{equation}
\Omega(\mathrm{OWCA}) = \left(1 + \frac{2}{r^2} \right) HWC^2 + (1 + \alpha)^2 \cdot \frac{2M^2}{r^2} HWC.
\end{equation}

This design balances efficiency and effectiveness by leveraging broader LR context for guiding the high-resolution refinement in CAL.

\subsection{Frequency domain L1 loss}
\label{subsec:FL1}

To explicitly supervise spectral fidelity, we propose a frequency-domain L1 loss ($\mathcal{L}_{\text{freq}}$) that penalizes discrepancies in both amplitude and phase of the Fourier spectrum. 
This aligns with our goal of mitigating aliasing and preserving fine details.

Let $\hat{F}$ and $F$ denote the discrete Fourier transforms of the predicted and ground-truth images, respectively, computed channel-wise:
\begin{equation}
\mathcal{L}_{\text{freq}} =
\frac{2}{UVC} \sum_{c=1}^{C} \sum_{u=0}^{\lfloor U/2 \rfloor}
\sum_{v=0}^{V-1}
\left| \hat{F}_c(u,v) - F_c(u,v) \right|_1,
\label{eq:freq_loss}
\end{equation}
where $(U, V)$ are the spectral dimensions and $|\cdot|_1$ denotes the sum of absolute differences over real and imaginary parts\footnote{
\resizebox{\columnwidth}{!}{
$\left| \hat{F}_c(u,v) - F_c(u,v) \right|_1 = 
\left| \operatorname{Re}\hat{F}_c(u,v) - \operatorname{Re}F_c(u,v) \right| +
\left| \operatorname{Im}\hat{F}_c(u,v) - \operatorname{Im}F_c(u,v) \right|$
}}.
The factor of $2$ compensates for Hermitian symmetry in real-valued images.

By constraining both amplitude and phase, this loss enforces fine-grained frequency consistency, reducing artifacts such as ringing, checkerboard effects, and spatial misalignment. 
It also remains computationally efficient since only the non-redundant half of the spectrum is evaluated.

%% file: figs/fga_architecture.tex
\begin{figure*}[!t]
    \centering
    \includegraphics[width=0.8\textwidth]{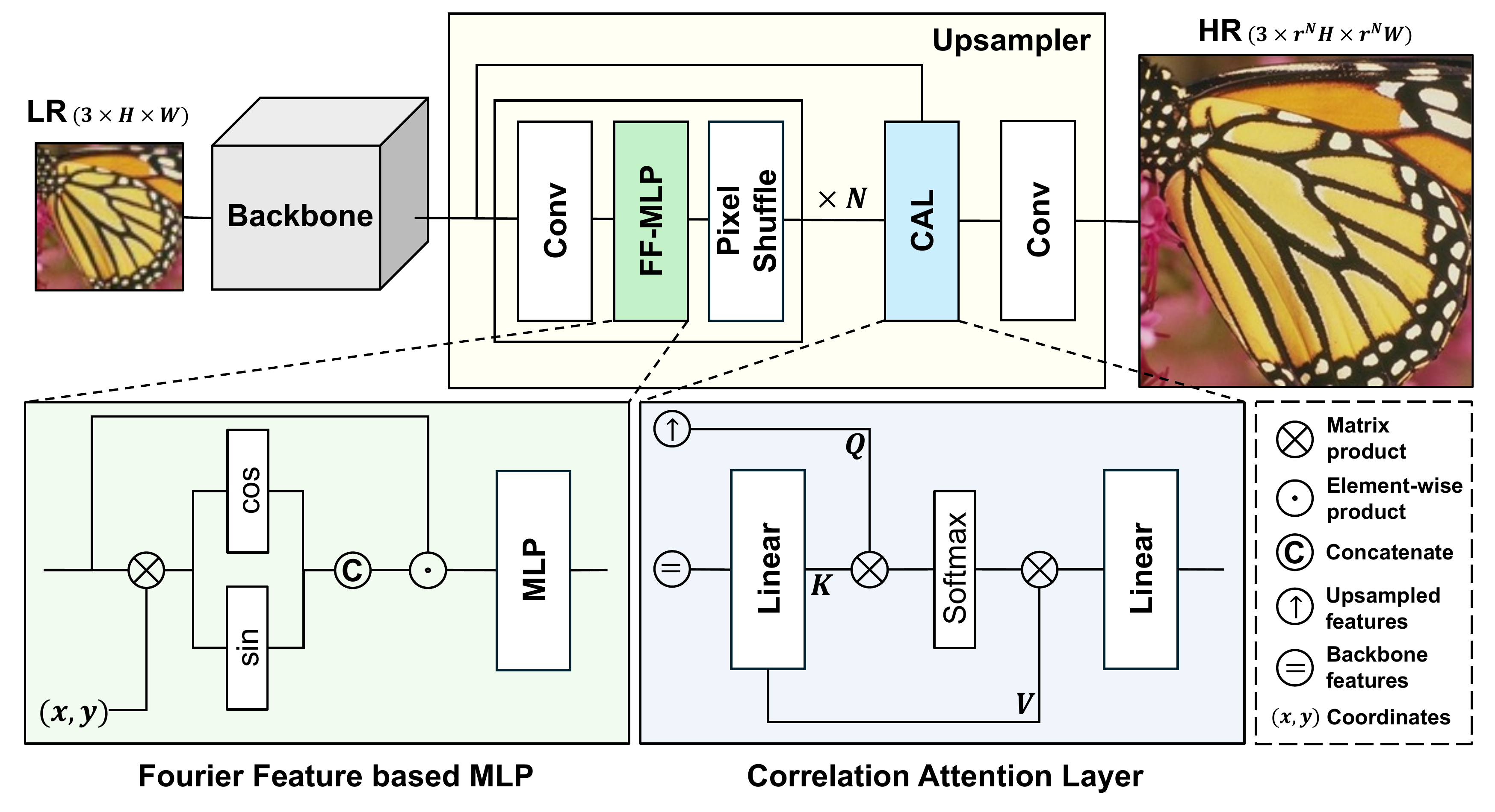}
    \caption{Overview of the proposed Fourier-Guided Attention (FGA) upsampler.
    }
    \label{fig:fga_architecture}
\end{figure*}

%% file: sec/04_experiment.tex
\newcommand{\lite}{\textsuperscript{$\mu$}}
\section{Experiments} \label{sec:experiment}

To assess the impact of our proposed FGA upsampler, we incorporate it into five widely adopted super-resolution backbones (EDSR~\cite{Conv-EDSR}, RCAN~\cite{ConvAttn-RCAN}, HAN~\cite{ConvAttn-HAN}, NLSN~\cite{ConvAttn-NLSN}, and SwinIR~\cite{SwinIR}) and benchmark each model on the standard $\times 4$ image super-resolution task.

Sub-Pixel Convolution (SPC) is used as the baseline upsampling module across all settings, enabling consistent comparison with our proposed FGA.
The proposed FGA module introduces approximately 0.3M additional parameters over SPC, and this overhead remains constant across all backbones, confirming that performance gains are not due to increased model capacity.
For the window-based cross-resolution attention in the FGA module (see Sec.~\ref{subsec:CAL}), we adopt a window size of $5 \times 5$ for pre-upsampling features and $4 \times 4$ for post-upsampling features, as this configuration provides the best trade-off between performance and efficiency based on empirical evaluation (see \ref{app:window_size}).

\paragraph{Two-Capacity Backbone Setting}
To isolate the effect of the upsampling module from overall model capacity, we evaluate each backbone under two distinct capacity settings (Tab.~\ref{tab:total_param}):
\begin{itemize}
  \item \textbf{Original variant:} Full-size models (12-40M parameters), initialized from publicly released checkpoints.
  \item \textbf{Lightweight variant (denoted as $^\mu$):} Reduced-capacity versions (1-9M parameters), obtained by halving both the channel width and depth, and trained from scratch.
\end{itemize}

\input{tables/params_results}

\paragraph{Training Protocol}
Unless otherwise stated, we follow the original training procedures for each backbone.
EDSR$^{\mu}$, RCAN$^{\mu}$, HAN$^{\mu}$, and NLSN$^{\mu}$ are trained on the DIV2K dataset~\cite{DIV2K} with a batch size of 16.
SwinIR$^{\mu}$ follows its original configuration and uses DF2K (DIV2K + Flickr2K~\cite{Flick2K}) with a batch size of 32.
All models are trained for 500K iterations on RTX A6000 GPUs, with learning rate decay applied at 250K, 400K, 450K, and 475K iterations. Evaluation is performed on a single RTX A6000 GPU.

\paragraph{Evaluation Datasets}
We evaluate model performance on six benchmark datasets: Set5~\cite{Set5}, Set14~\cite{Set14}, BSD100~\cite{BSDS100}, Urban100~\cite{Urban100}, Manga109~\cite{Manga109}, and DTD235.
DTD235 is a custom subset of 235 images randomly selected from the Describable Textures Dataset (DTD~\cite{DTD}), curated to emphasize challenging high-frequency textures.

\subsection{Results}\label{subsec:result}
\input{tables/quantitative_results}
Tab.~\ref{tab:quantitative_results} presents PSNR and SSIM scores on six benchmarks for $\times4$ super-resolution using five backbone models under both original and lightweight (\textsuperscript{$\mu$}) configurations.
For every setting, we compare the baseline SPC upsampler against our proposed FGA module.

FGA consistently outperforms SPC across all architectures and datasets, confirming its effectiveness in enhancing reconstruction quality without significantly increasing model complexity.
On average, FGA yields a PSNR improvement of +0.14 dB for lightweight networks and +0.12 dB for full-capacity (i.e., original) models.
The gains are especially prominent on high-frequency benchmarks such as Urban100 and Manga109, where fine texture recovery is critical.
For instance, replacing SPC with FGA in NLSN$^{\mu}$ improves PSNR from 26.37 dB to 26.64 dB on Urban100 (+0.27 dB) and from 30.75 dB to 31.17 dB on Manga109 (+0.42 dB).

These results highlight that the benefits of FGA scale reliably with backbone size, even under the fixed additional cost of only 0.3M parameters.
The consistent improvements in both PSNR and SSIM validate that our frequency-aware upsampling strategy effectively enhances structural fidelity while reducing artifacts, irrespective of the underlying network capacity.

\paragraph{Qualitative Comparison}

\input{figs/sr_results}

Figure~\ref{fig:sr_results} provides qualitative comparisons across datasets and backbones under $\times$4 upsampling.
Across all backbones and datasets, FGA visibly improves the reconstruction of fine details such as repetitive structures, sharp contours, and textures, while reducing aliasing artifacts that are often present with SPC upsampler.
These improvements are evident across both lightweight and full-capacity backbones, reinforcing that FGA effectively enhances spatial detail and reduces aliasing regardless of model size.

\subsection{Spectral Feature Analysis}
\label{subsec:spectral_analysis}

To better understand the source of the improvements offered by FGA, we revisit the frequency-space characteristics of feature maps produced during super-resolution.
Following the analysis in Section~\ref{sec:analysis_artifacts}, we examine intermediate representations both before and after upsampling, with a particular focus on how our FGA modifies the spectral structure compared to SPC.

Fig.~\ref{fig:feat_results_edsr} shows PCA-reduced feature maps and frequency spectra from EDSR using three different upsampling strategies.
The spectral maps indicate that SPC and deconvolution layers suffer from visible aliasing, including replicated high-frequency patterns and checkerboard artifacts.
In contrast, FGA suppresses these patterns, yielding frequency spectra that more closely align with the GT distributions.
These observations suggest that FGA successfully mitigates spectral folding and enhances local structure.

To confirm that these improvements generalize across architectures, Fig.~\ref{fig:feat_results_swinir} presents the same analysis on SwinIR$^\mu$.
Even on unseen Urban100 test images, the post-upsampling frequency maps from FGA retain higher-frequency components and exhibit fewer aliasing artifacts than SPC.
The visualizations in Fig.~\ref{fig:feat_map_swinir} also reveal that the spatial feature maps generated by FGA are less noisy and exhibit clearer edge transitions.

Taken together, these results provide strong visual evidence that the proposed FGA reduces spectral aliasing directly at the feature level.
By explicitly encoding spatial frequency and positional information prior to upsampling, FGA promotes more coherent feature transformations, which translate into perceptually sharper and structurally consistent reconstructions.

\subsection{Quantitative Frequency Analysis via FRC}
\label{subsec:frc}

\input{figs/frc_graphs_and_ring}

To quantitatively evaluate the frequency-domain consistency of SR outputs, we adopt the Fourier Ring Correlation (FRC), originally developed for resolution estimation in cryo-EM~\cite{FRC_0,FRC_1}.
FRC measures structural similarity between two images in the Fourier domain by computing correlation along concentric rings in frequency space.
Given 2D Fourier transforms $F_1(\mathbf{f})$ and $F_2(\mathbf{f})$ of two images (e.g., SR output and GT), the FRC score at spatial frequency radius $q = \lVert \mathbf{f} \rVert$ is defined as:
\begin{equation}
\small
\mathrm{FRC}(q)=
\frac{\displaystyle\sum_{|\mathbf f|=q}F_{1}(\mathbf f)\,F_{2}^{*}(\mathbf f)}
     {\displaystyle\sqrt{\Bigl(\sum_{|\mathbf f|=q}|F_{1}(\mathbf f)|^{2}\Bigr)
                         \Bigl(\sum_{|\mathbf f|=q}|F_{2}(\mathbf f)|^{2}\Bigr)}}
\label{eq:frc}
\end{equation}
Following prior work~\cite{FRC_2}, we quantize the 2D frequency domain into $N = 64$ concentric rings, each containing approximately equal numbers of frequency-domain pixels (see Fig.~\ref{fig:frc_ring}), enabling fine-grained, frequency-wise comparisons between the reconstructed and reference images.

\paragraph{Top-Quartile FRC for Fine Detail}
Across all tested datasets, the first 75\% of the FRC curve remains close to 1.0, reflecting strong agreement on low-to-mid frequencies that dominate global structure (e.g., Fig.~\ref{fig:frc_graph_set5} and \ref{fig:frc_graph_urban100}).
However, meaningful differences emerge in the top 25\% of rings (48–63), which correspond to high-frequency content such as textures and fine details.
We define the average FRC in this upper quartile as a scalar metric:
\begin{equation}
\mathrm{FRC\text{-}AUC} =
\frac{1}{N_{\text{HF}}}
\sum_{i=i_{\mathrm{HF}}}^{N - 1} \mathrm{FRC}_i,
\quad i_{\mathrm{HF}} = \left\lceil 0.75N \right\rceil = 48,
\label{eq:hffrcauc}
\end{equation}
where $N_{\text{HF}} = 16$ ($N - i_{\mathrm{HF}}$).
This metric provides a concise and targeted measure of fine-detail fidelity in SR reconstructions.

\paragraph{FRC-AUC Results Analysis}
\input{tables/frc_auc_results}

Tab.~\ref{tab:frc_auc_results} reports the FRC-AUC scores across six benchmarks for all evaluated backbones and both upsamplers (SPC vs. FGA).
Across the board, replacing SPC with our proposed FGA consistently improves high-frequency fidelity. 
Gains are especially prominent on texture-rich datasets such as Urban100 and Manga109.
For example, NLSN$^\mu$ improves from 0.2330 to 0.3017 on Urban100 (+29\%), while EDSR$^\mu$ and SwinIR$^\mu$ achieve similar relative gains (+26\%).
Even in full-capacity models, improvements are non-trivial, for instance, NLSN rises from 0.3019 to 0.3178 (+5.3\%).
Interestingly, we observe that HAN achieves a lower FRC-AUC score than its lightweight counterpart HAN$^\mu$, which we attribute to halo artifacts around edges in some samples (see \ref{app:han}).
Nevertheless, our proposed FGA consistently improves the FRC-AUC scores for both HAN and HAN$^\mu$, demonstrating its robustness across backbone variants.

On average, FGA yields a +10\% increase in FRC-AUC on Urban100 and +7\% on Manga109 compared to SPC, demonstrating its superior preservation of high-frequency content.
Interestingly, lightweight variants using FGA often outperform larger SPC-based counterparts, for example, EDSR$^\mu$ + FGA surpasses full EDSR + SPC on Set5, Manga109, and DTD235.
Similar cross-scale improvements appear in RCAN and SwinIR as well.

These results confirm that the frequency-aware design of FGA offers consistent improvements in spectral detail across architectures and model capacities, validating its effectiveness as a drop-in upsampling replacement.

\paragraph{Visual Comparison of High-Frequency Bands}
\input{figs/hf_vis_results}
We further visualize the highest-frequency quartile in Fig.~\ref{fig:hf_vis_results}.
These maps isolate the detail regions where SR models differ most.
Across all backbones, FGA produces denser, more coherent high-frequency patterns that align closely with the GT reference, while SPC outputs appear fragmented or noisy.
Zoom-in views reveal that FGA maintains subtle periodicity and preserves clean edge transitions, whereas SPC often introduces residual ringing and scattered noise.
These patterns confirm that our Fourier-guided design strengthens detail fidelity across architectures, despite adding only a marginal number of parameters.

\subsection{Ablation Study}\label{subsec:ablation}
We conduct an ablation study to quantify the contribution of each component in our FGA upsampler, including MLP-based channel mixing (MLP), Fourier-feature positional encoding (FF), correlation attention layer (CAL), and the frequency-domain L1 loss (FL1).
Variants are created by selectively disabling modules or replacing FL1 with the standard pixel-domain L1 loss.
\input{tables/ablation_study}
Results are summarized in Tab.~\ref{tab:ablation_study}.
The complete configuration (MLP + FF + CA trained with FL1) achieves the highest performance, attaining 32.30 dB PSNR and 0.8966 SSIM on Set5, along with a FRC-AUC of 0.2934.
On Urban100, it reaches 26.19 dB PSNR, 0.7867 SSIM, and 0.2743 FRC-AUC
Removing either FF or CAL leads to noticeable degradation: PSNR drops by 0.1–0.2 dB and FRC-AUC declines by 4–9\%.
Excluding both components or substituting FL1 with L1 results in the most substantial performance drops across all metrics.

Interestingly, adding MLP alone without FF yields limited benefits—its PSNR (32.11 dB) and FRC-AUC (0.2673) are both lower than when FF is included—suggesting that channel mixing alone cannot recover spatial-frequency cues.
Overall, FF provides essential position-aware spectral context, enabling CAL and MLP to function effectively, while FL1 further promotes high-frequency fidelity.
These results validate the complementary design of the proposed components.

%% file: tables/params_results.tex
\begin{table}[h]
\centering
\caption{Parameter count (in millions) for each backbone using either SPC or our proposed FGA upsampler. Each model is evaluated in both lightweight ($^{\mu}$) and original variants.}
\footnotesize
\begin{tabular}{lcccccc}
\toprule
\textbf{Setting} & \textbf{EDSR} & \textbf{RCAN} & \textbf{HAN} & \textbf{NLSN} & \textbf{SwinIR} \\
\midrule
$^{\mu}$ + SPC & 1.5  & 4.3  & 8.6  & 1.6  & 1.2 \\
$^{\mu}$ + FGA & 1.8  & 4.6  & 8.9  & 1.9  & 1.5 \\
Original + SPC & 38.7 & 15.6 & 16.1 & 39.7 & 11.9 \\
Original + FGA & 39.0 & 15.9 & 16.4 & 40.0 & 12.2 \\
\bottomrule
\end{tabular}
\label{tab:total_param}
\end{table}

%% file: tables/quantitative_results.tex
\begin{table*}[!th]
  \centering
  \caption{
  Quantitative $\times$4 SR results on six benchmark datasets, reported in terms of PSNR (dB) / SSIM.
  Each backbone is evaluated in both its original and lightweight variant (denoted as “$^\mu$”).
  For each configuration, the top row uses SPC as the baseline upsampler, while the bottom row uses our proposed FGA module.
  Bold values indicate superior performance between SPC and FGA for each setting.
  }
  \resizebox{\textwidth}{!}{%
    \begin{tabular}{ll cccccccccccc}
    \toprule
    \multicolumn{1}{c}{} & & 
      \multicolumn{2}{c}{\textbf{Set5}} &
      \multicolumn{2}{c}{\textbf{Set14}} &
      \multicolumn{2}{c}{\textbf{BSD100}} &
      \multicolumn{2}{c}{\textbf{Urban100}} &
      \multicolumn{2}{c}{\textbf{Manga109}} &
      \multicolumn{2}{c}{\textbf{DTD235}} \\
    \cmidrule(lr){3-4}\cmidrule(lr){5-6}\cmidrule(lr){7-8}\cmidrule(lr){9-10}\cmidrule(lr){11-12}\cmidrule(lr){13-14}
      \multirow{-2}{*}[2pt]{\textbf{Backbone}} &
      \multirow{-2}{*}[2pt]{\textbf{Upsampler}} &
      \multicolumn{1}{c}{\textbf{PSNR}} &
      \multicolumn{1}{c}{\textbf{SSIM}} &
      \multicolumn{1}{c}{\textbf{PSNR}} &
      \multicolumn{1}{c}{\textbf{SSIM}} &
      \multicolumn{1}{c}{\textbf{PSNR}} &
      \multicolumn{1}{c}{\textbf{SSIM}} &
      \multicolumn{1}{c}{\textbf{PSNR}} &
      \multicolumn{1}{c}{\textbf{SSIM}} &
      \multicolumn{1}{c}{\textbf{PSNR}} &
      \multicolumn{1}{c}{\textbf{SSIM}} &
      \multicolumn{1}{c}{\textbf{PSNR}} &
      \multicolumn{1}{c}{\textbf{SSIM}} \\
    \midrule
    \midrule
    \multicolumn{1}{l}{} & SPC & 32.12 & 0.8953 & 28.59 & 0.7822 & 27.58 & 0.7378 & 26.08 & 0.7860 & 30.41 & 0.9077 & 29.73 & 0.7599 \\
    \multicolumn{1}{l}{\multirow{-2}{*}{EDSR\textsuperscript{$\mu$}}} & FGA \small{(ours)} & \textbf{32.30} & \textbf{0.8968} & \textbf{28.68} & \textbf{0.7838} & \textbf{27.64} & \textbf{0.7392} & \textbf{26.27} & \textbf{0.7896} & \textbf{30.74} & \textbf{0.9106} & \textbf{29.80} & \textbf{0.7601} \\
    \midrule
    \multicolumn{1}{l}{} & SPC & 32.46 & 0.8968 & 28.80 & 0.7876 & 27.71 & 0.7420 & 26.64 & \textbf{0.8033} & 31.02 & 0.9148 & \textbf{29.91} & \textbf{0.7664} \\
    \multicolumn{1}{l}{\multirow{-2}{*}{EDSR}} & FGA \small{(ours)} & \textbf{32.50} & \textbf{0.8995} & 28.80 & \textbf{0.7877} & \textbf{27.74} & \textbf{0.7437} & \textbf{26.67} & 0.8030 & \textbf{31.05} & \textbf{0.9158} & 29.89 & 0.7655 \\
    \midrule
    \multicolumn{1}{l}{} & SPC & 32.34 & 0.8975 & 28.69 & 0.7851 & 27.65 & 0.7406 & 26.44 & 0.7972 & 30.78 & 0.9124 & 29.88 & \textbf{0.7649} \\
    \multicolumn{1}{l}{\multirow{-2}{*}{RCAN\textsuperscript{$\mu$}}} & FGA \small{(ours)} & \textbf{32.43} & \textbf{0.8984} & \textbf{28.76} & \textbf{0.7858} & \textbf{27.70} & \textbf{0.7414} & \textbf{26.52} & \textbf{0.7974} & \textbf{30.94} & \textbf{0.9135} & \textbf{29.90} & 0.7641 \\
    \midrule
    \multicolumn{1}{l}{} & SPC & 32.63 & 0.9002 & 28.87 & 0.7889 & 27.77 & 0.7436 & 26.82 & \textbf{0.8087} & 31.22 & \textbf{0.9173} & 30.00 & \textbf{0.7681} \\
    \multicolumn{1}{l}{\multirow{-2}{*}{RCAN}} & FGA \small{(ours)} & 32.63 & \textbf{0.9007} & \textbf{28.88} & \textbf{0.7891} & \textbf{27.79} & \textbf{0.7449} & \textbf{26.83} & 0.8067 & \textbf{31.23} & 0.9161 & \textbf{30.02} & 0.7678 \\
    \midrule
    \multicolumn{1}{l}{} & SPC & 32.44 & 0.8988 & 28.76 & 0.7865 & 27.70 & 0.7419 & 26.55 & 0.8002 & 30.96 & 0.9142 & 29.92 & \textbf{0.7665} \\
    \multicolumn{1}{l}{\multirow{-2}{*}{HAN\textsuperscript{$\mu$}}}  & FGA \small{(ours)} & \textbf{32.52} & \textbf{0.8997} & \textbf{28.85} & \textbf{0.7878} & \textbf{27.75} & \textbf{0.7431} & \textbf{26.70} & \textbf{0.8024} & \textbf{31.21} & \textbf{0.9160} & \textbf{29.97} & 0.7660 \\
    \midrule
    \multicolumn{1}{l}{} & SPC & 32.64 & 0.9002 & 28.90 & 0.7890 & 27.80 & 0.7442 & 26.85 & \textbf{0.8094} & 31.42 & 0.9177 & 30.06 & \textbf{0.7703} \\
    \multicolumn{1}{l}{\multirow{-2}{*}{HAN}}  & FGA \small{(ours)} & \textbf{32.68} & \textbf{0.9013} & \textbf{28.93} & \textbf{0.7897} & \textbf{27.81} & \textbf{0.7454} & \textbf{26.91} & 0.8087 & \textbf{31.43} & \textbf{0.9181} & 30.06 & 0.7689 \\
    \midrule
    \multicolumn{1}{l}{} & SPC & 32.30 & 0.8969 & 28.69 & 0.7841 & 27.65 & 0.7401 & 26.37 & 0.7937 & 30.75 & 0.9112 & 29.84 & 0.7653 \\
    \multicolumn{1}{l}{\multirow{-2}{*}{NLSN\textsuperscript{$\mu$}}} & FGA \small{(ours)} & \textbf{32.48} & \textbf{0.8990} & \textbf{28.81} & \textbf{0.7863} & \textbf{27.73} & \textbf{0.7425} & \textbf{26.64} & \textbf{0.7994} & \textbf{31.17} & \textbf{0.9155} & \textbf{29.96} & \textbf{0.7666} \\
    \midrule
    \multicolumn{1}{l}{} & SPC & 32.59 & 0.9000 & 28.87 & 0.7891 & 27.78 & 0.7444 & 26.96 & 0.8109 & 31.27 & 0.9184 & 30.08 & \textbf{0.7715} \\
    \multicolumn{1}{l}{\multirow{-2}{*}{NLSN}} & FGA \small{(ours)} & \textbf{32.72} & \textbf{0.9021} & \textbf{28.98} & \textbf{0.7907} & \textbf{27.85} & \textbf{0.7471} & \textbf{27.09} & \textbf{0.8136} & \textbf{31.59} & \textbf{0.9207} & \textbf{30.11} & 0.7713 \\
    \midrule
    \multicolumn{1}{l}{} & SPC & 32.49 & 0.8994 & 28.86 & 0.7884 & 27.75 & 0.7441 & 26.66 & 0.8040 & 31.23 & 0.9172 & 29.96 & 0.7683 \\
    \multicolumn{1}{l}{\multirow{-2}{*}{SwinIR\textsuperscript{$\mu$}}} & FGA \small{(ours)} & \textbf{32.59} & \textbf{0.9007} & \textbf{28.97} & \textbf{0.7905} & \textbf{27.82} & \textbf{0.7455} & \textbf{26.86} & \textbf{0.8064} & \textbf{31.55} & \textbf{0.9195} & \textbf{30.06} & \textbf{0.7693} \\
    \midrule
    \multicolumn{1}{l}{} & SPC & 32.92 & 0.9044 & 29.09 & 0.7950 & 27.92 & 0.7489 & 27.45 & \textbf{0.8254} & 32.03 & 0.9260 & 30.28 & \textbf{0.7789} \\
    \multicolumn{1}{l}{\multirow{-2}{*}{SwinIR}} & FGA \small{(ours)} & \textbf{32.96} & \textbf{0.9052} & \textbf{29.13} & \textbf{0.7957} & \textbf{27.96} & \textbf{0.7510} & \textbf{27.47} & 0.8249 & \textbf{32.14} & \textbf{0.9261} & \textbf{30.29} & 0.7774 \\
    \bottomrule
    \end{tabular}%
  }
  \label{tab:quantitative_results}
\end{table*}

%% file: figs/sr_results.tex
\begin{figure*}[t]
    \renewcommand{\arraystretch}{1.1}
    \renewcommand{\tabcolsep}{0.9mm}
    \resizebox{\textwidth}{!}{%
        \begin{tabular}{cc ccccc cc c ccccc}
            & & \small LW-scale & \small Origin-scale & & \small Cropped img & \small img\_072 
            & \multicolumn{1}{c}{}
            & \multicolumn{1}{c}{} &
            & \small LW-scale & \small Origin-scale & & \small Cropped img & \small BokuHaSitatakaKun \\
            \multirow{2}{*}[12pt]{\rotatebox{90}{\textbf{(a) EDSR}}}
            & \multirow{1}{*}[35pt]{\rotatebox{90}{\small SPC}}
            & \includegraphics[height=0.15\linewidth, width=0.15\linewidth]{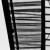}
            & \includegraphics[height=0.15\linewidth, width=0.15\linewidth]{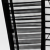} \rule[-5pt]{1pt}{80pt}
            & \multirow{1}{*}[35pt]{\rotatebox{90}{\small HR}}
            & \includegraphics[height=0.15\linewidth, width=0.15\linewidth]{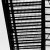}
            & \multirow{2}{*}[65pt]{\includegraphics[height=0.31\linewidth, width=0.31\linewidth]{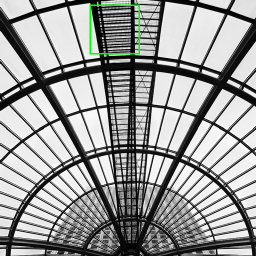}}
            & \multicolumn{1}{c}{}
            & \multicolumn{1}{c}{}
            & \multirow{1}{*}[35pt]{\rotatebox{90}{\small SPC}}
            & \includegraphics[height=0.15\linewidth, width=0.15\linewidth]{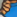}
            & \includegraphics[height=0.15\linewidth, width=0.15\linewidth]{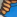} \rule[-5pt]{1pt}{80pt}
            & \multirow{1}{*}[35pt]{\rotatebox{90}{\small HR}}
            & \includegraphics[height=0.15\linewidth, width=0.15\linewidth]{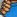}
            & \multirow{2}{*}[65pt]{\includegraphics[height=0.31\linewidth, width=0.31\linewidth]{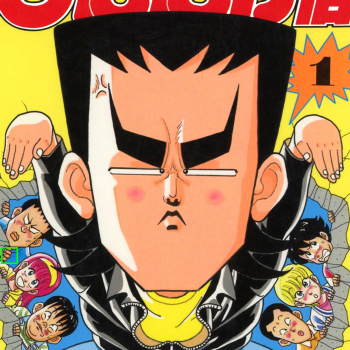}}
            \\
            
            & \multirow{1}{*}[35pt]{\rotatebox{90}{\small FGA}}
            & \includegraphics[height=0.15\linewidth, width=0.15\linewidth]{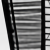}
            & \includegraphics[height=0.15\linewidth, width=0.15\linewidth]{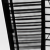} \rule[0pt]{1pt}{75pt}
            & \multirow{1}{*}[35pt]{\rotatebox{90}{\small LR}}
            & \includegraphics[height=0.15\linewidth, width=0.15\linewidth]{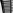}
            &
            & \multicolumn{1}{c}{}
            & \multicolumn{1}{c}{}
            & \multirow{1}{*}[35pt]{\rotatebox{90}{\small FGA}}
            & \includegraphics[height=0.15\linewidth, width=0.15\linewidth]{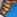}
            & \includegraphics[height=0.15\linewidth, width=0.15\linewidth]{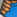} \rule[0pt]{1pt}{75pt}
            & \multirow{1}{*}[35pt]{\rotatebox{90}{\small LR}}
            & \includegraphics[height=0.15\linewidth, width=0.15\linewidth]{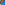}
            & 
            \\ [15pt]
            
            & & \small LW-scale & \small Origin-scale & & \small Cropped img & \small img\_024 
            & \multicolumn{1}{c}{}
            & \multicolumn{1}{c}{} & 
            & \small LW-scale & \small Origin-scale & & \small Cropped img & \small grid\_0025 \\
            \multirow{2}{*}[12pt]{\rotatebox{90}{\textbf{(b) RCAN}}}
            & \multirow{1}{*}[35pt]{\rotatebox{90}{\small SPC}}
            & \includegraphics[height=0.15\linewidth, width=0.15\linewidth]{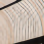}
            & \includegraphics[height=0.15\linewidth, width=0.15\linewidth]{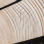} \rule[-5pt]{1pt}{80pt}
            & \multirow{1}{*}[35pt]{\rotatebox{90}{\small HR}}
            & \includegraphics[height=0.15\linewidth, width=0.15\linewidth]{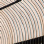}
            & \multirow{2}{*}[65pt]{\includegraphics[height=0.31\linewidth, width=0.31\linewidth]{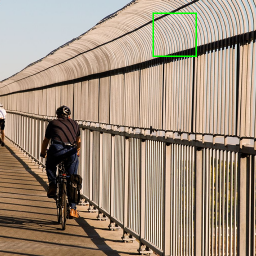}}
            & \multicolumn{1}{c}{}
            & \multicolumn{1}{c}{}
            & \multirow{1}{*}[35pt]{\rotatebox{90}{\small SPC}}
            & \includegraphics[height=0.15\linewidth, width=0.15\linewidth]{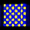}
            & \includegraphics[height=0.15\linewidth, width=0.15\linewidth]{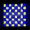} \rule[-5pt]{1pt}{80pt}
            & \multirow{1}{*}[35pt]{\rotatebox{90}{\small HR}}
            & \includegraphics[height=0.15\linewidth, width=0.15\linewidth]{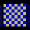}
            & \multirow{2}{*}[65pt]{\includegraphics[height=0.31\linewidth, width=0.31\linewidth]{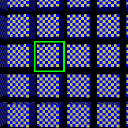}}
            \\
            
            & \multirow{1}{*}[35pt]{\rotatebox{90}{\small FGA}}
            & \includegraphics[height=0.15\linewidth, width=0.15\linewidth]{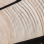}
            & \includegraphics[height=0.15\linewidth, width=0.15\linewidth]{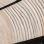} \rule[0pt]{1pt}{75pt}
            & \multirow{1}{*}[35pt]{\rotatebox{90}{\small LR}}
            & \includegraphics[height=0.15\linewidth, width=0.15\linewidth]{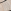}
            & 
            & \multicolumn{1}{c}{}
            & \multicolumn{1}{c}{}
            & \multirow{1}{*}[35pt]{\rotatebox{90}{\small FGA}}
            & \includegraphics[height=0.15\linewidth, width=0.15\linewidth]{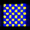}
            & \includegraphics[height=0.15\linewidth, width=0.15\linewidth]{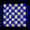} \rule[0pt]{1pt}{75pt}
            & \multirow{1}{*}[35pt]{\rotatebox{90}{\small LR}}
            & \includegraphics[height=0.15\linewidth, width=0.15\linewidth]{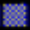}
            & 
            \\ [15pt]

            & & \small LW-scale & \small Origin-scale & & \small Cropped img & \small img\_092 
            & \multicolumn{1}{c}{}
            & \multicolumn{1}{c}{} & 
            & \small LW-scale & \small Origin-scale & & \small Cropped img & \small YasasiiAkuma \\
            \multirow{2}{*}[12pt]{\rotatebox{90}{\textbf{(c) HAN}}}
            & \multirow{1}{*}[35pt]{\rotatebox{90}{\small SPC}}
            & \includegraphics[height=0.15\linewidth, width=0.15\linewidth]{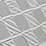}
            & \includegraphics[height=0.15\linewidth, width=0.15\linewidth]{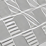} \rule[-5pt]{1pt}{80pt}
            & \multirow{1}{*}[35pt]{\rotatebox{90}{\small HR}}
            & \includegraphics[height=0.15\linewidth, width=0.15\linewidth]{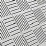}
            & \multirow{2}{*}[65pt]{\includegraphics[height=0.31\linewidth, width=0.31\linewidth]{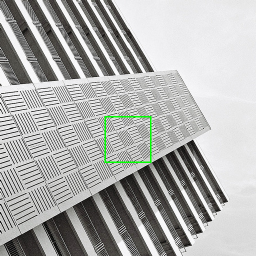}}
            & \multicolumn{1}{c}{}
            & \multicolumn{1}{c}{}
            & \multirow{1}{*}[35pt]{\rotatebox{90}{\small SPC}}
            & \includegraphics[height=0.15\linewidth, width=0.15\linewidth]{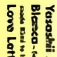}
            & \includegraphics[height=0.15\linewidth, width=0.15\linewidth]{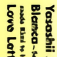} \rule[-5pt]{1pt}{80pt}
            & \multirow{1}{*}[35pt]{\rotatebox{90}{\small HR}}
            & \includegraphics[height=0.15\linewidth, width=0.15\linewidth]{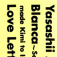}
            & \multirow{2}{*}[65pt]{\includegraphics[height=0.31\linewidth, width=0.31\linewidth]{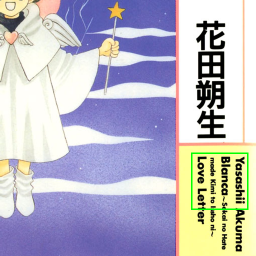}}
            \\
            
            & \multirow{1}{*}[35pt]{\rotatebox{90}{\small FGA}}
            & \includegraphics[height=0.15\linewidth, width=0.15\linewidth]{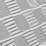}
            & \includegraphics[height=0.15\linewidth, width=0.15\linewidth]{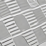} \rule[0pt]{1pt}{75pt}
            & \multirow{1}{*}[35pt]{\rotatebox{90}{\small LR}}
            & \includegraphics[height=0.15\linewidth, width=0.15\linewidth]{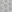}
            & 
            & \multicolumn{1}{c}{}
            & \multicolumn{1}{c}{}
            & \multirow{1}{*}[35pt]{\rotatebox{90}{\small FGA}}
            & \includegraphics[height=0.15\linewidth, width=0.15\linewidth]{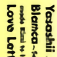}
            & \includegraphics[height=0.15\linewidth, width=0.15\linewidth]{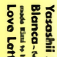} \rule[0pt]{1pt}{75pt}
            & \multirow{1}{*}[35pt]{\rotatebox{90}{\small LR}}
            & \includegraphics[height=0.15\linewidth, width=0.15\linewidth]{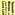}
            & 
            \\ [15pt]

            & & \small LW-scale & \small Origin-scale & & \small Cropped img & \small img\_083
            & \multicolumn{1}{c}{}
            & \multicolumn{1}{c}{} & 
            & \small LW-scale & \small Origin-scale & & \small Cropped img & \small PikaruGenkiDesu \\
            \multirow{2}{*}[12pt]{\rotatebox{90}{\textbf{(d) NLSN}}}
            & \multirow{1}{*}[35pt]{\rotatebox{90}{\small SPC}}
            & \includegraphics[height=0.15\linewidth, width=0.15\linewidth]{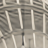}
            & \includegraphics[height=0.15\linewidth, width=0.15\linewidth]{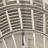} \rule[-5pt]{1pt}{80pt}
            & \multirow{1}{*}[35pt]{\rotatebox{90}{\small HR}}
            & \includegraphics[height=0.15\linewidth, width=0.15\linewidth]{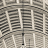}
            & \multirow{2}{*}[65pt]{\includegraphics[height=0.31\linewidth, width=0.31\linewidth]{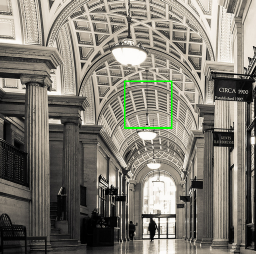}}
            & \multicolumn{1}{c}{}
            & \multicolumn{1}{c}{}
            & \multirow{1}{*}[35pt]{\rotatebox{90}{\small SPC}}
            & \includegraphics[height=0.15\linewidth, width=0.15\linewidth]{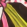}
            & \includegraphics[height=0.15\linewidth, width=0.15\linewidth]{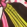} \rule[-5pt]{1pt}{80pt}
            & \multirow{1}{*}[35pt]{\rotatebox{90}{\small HR}}
            & \includegraphics[height=0.15\linewidth, width=0.15\linewidth]{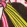}
            & \multirow{2}{*}[65pt]{\includegraphics[height=0.31\linewidth, width=0.31\linewidth]{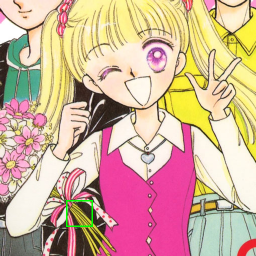}}
            \\
            
            & \multirow{1}{*}[35pt]{\rotatebox{90}{\small FGA}}
            & \includegraphics[height=0.15\linewidth, width=0.15\linewidth]{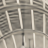}
            & \includegraphics[height=0.15\linewidth, width=0.15\linewidth]{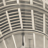} \rule[0pt]{1pt}{75pt}
            & \multirow{1}{*}[35pt]{\rotatebox{90}{\small LR}}
            & \includegraphics[height=0.15\linewidth, width=0.15\linewidth]{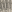}
            & 
            & \multicolumn{1}{c}{}
            & \multicolumn{1}{c}{}
            & \multirow{1}{*}[35pt]{\rotatebox{90}{\small FGA}}
            & \includegraphics[height=0.15\linewidth, width=0.15\linewidth]{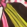}
            & \includegraphics[height=0.15\linewidth, width=0.15\linewidth]{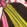} \rule[0pt]{1pt}{75pt}
            & \multirow{1}{*}[35pt]{\rotatebox{90}{\small LR}}
            & \includegraphics[height=0.15\linewidth, width=0.15\linewidth]{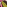}
            & 
            \\ [15pt]

            & & \small LW-scale & \small Origin-scale & & \small Cropped img & \small img\_078 
            & \multicolumn{1}{c}{}
            & \multicolumn{1}{c}{} & 
            & \small LW-scale & \small Origin-scale & & \small Cropped img & \small UchuKigekiM774 \\
            \multirow{2}{*}[12pt]{\rotatebox{90}{\textbf{(e) SwinIR}}}
            & \multirow{1}{*}[35pt]{\rotatebox{90}{\small SPC}}
            & \includegraphics[height=0.15\linewidth, width=0.15\linewidth]{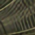}
            & \includegraphics[height=0.15\linewidth, width=0.15\linewidth]{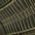} \rule[-5pt]{1pt}{80pt}
            & \multirow{1}{*}[35pt]{\rotatebox{90}{\small HR}}
            & \includegraphics[height=0.15\linewidth, width=0.15\linewidth]{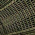}
            & \multirow{2}{*}[65pt]{\includegraphics[height=0.31\linewidth, width=0.31\linewidth]{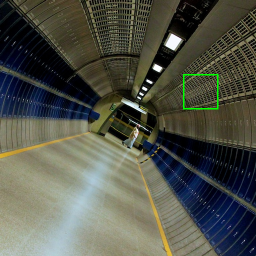}}
            & \multicolumn{1}{c}{}
            & \multicolumn{1}{c}{}
            & \multirow{1}{*}[35pt]{\rotatebox{90}{\small SPC}}
            & \includegraphics[height=0.15\linewidth, width=0.15\linewidth]{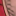}
            & \includegraphics[height=0.15\linewidth, width=0.15\linewidth]{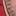} \rule[-5pt]{1pt}{80pt}
            & \multirow{1}{*}[35pt]{\rotatebox{90}{\small HR}}
            & \includegraphics[height=0.15\linewidth, width=0.15\linewidth]{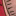}
            & \multirow{2}{*}[65pt]{\includegraphics[height=0.31\linewidth, width=0.31\linewidth]{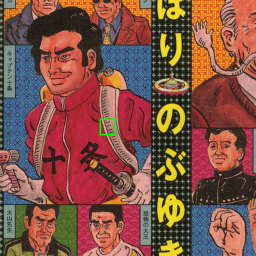}}
            \\
            
            & \multirow{1}{*}[35pt]{\rotatebox{90}{\small FGA}}
            & \includegraphics[height=0.15\linewidth, width=0.15\linewidth]{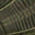}
            & \includegraphics[height=0.15\linewidth, width=0.15\linewidth]{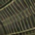} \rule[0pt]{1pt}{75pt}
            & \multirow{1}{*}[35pt]{\rotatebox{90}{\small LR}}
            & \includegraphics[height=0.15\linewidth, width=0.15\linewidth]{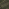}
            & 
            & \multicolumn{1}{c}{}
            & \multicolumn{1}{c}{}
            & \multirow{1}{*}[35pt]{\rotatebox{90}{\small FGA}}
            & \includegraphics[height=0.15\linewidth, width=0.15\linewidth]{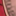}
            & \includegraphics[height=0.15\linewidth, width=0.15\linewidth]{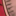} \rule[0pt]{1pt}{75pt}
            & \multirow{1}{*}[35pt]{\rotatebox{90}{\small LR}}
            & \includegraphics[height=0.15\linewidth, width=0.15\linewidth]{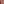}
            & 
            \end{tabular}
        }
    \caption{Qualitative $\times4$ SR results on Urban100, Manga109, and DTD235~\cite{Urban100, Manga109, DTD}.
            Each row compares SPC (top) with our FGA (bottom) across lightweight (LW-scale) and original (Origin-scale) variants of five SR backbones~\cite{Conv-EDSR, ConvAttn-RCAN, ConvAttn-HAN, ConvAttn-NLSN, SwinIR}.
            The rightmost column shows the full HR reference; green boxes indicate zoomed-in regions.
    }
    \label{fig:sr_results}
\end{figure*} 

%% file: figs/frc_graphs_and_ring.tex
\begin{figure*}[!t]
  \centering
    \resizebox{0.8\linewidth}{!}{%
    \begin{tabular}{cc}
      \phantomsubcaption\label{fig:frc_graph_set5}
      (a) Average FRC curve on Set5 &
      (c) Equal-pixel ring indexing map \\

      \includegraphics[width=0.5\linewidth]{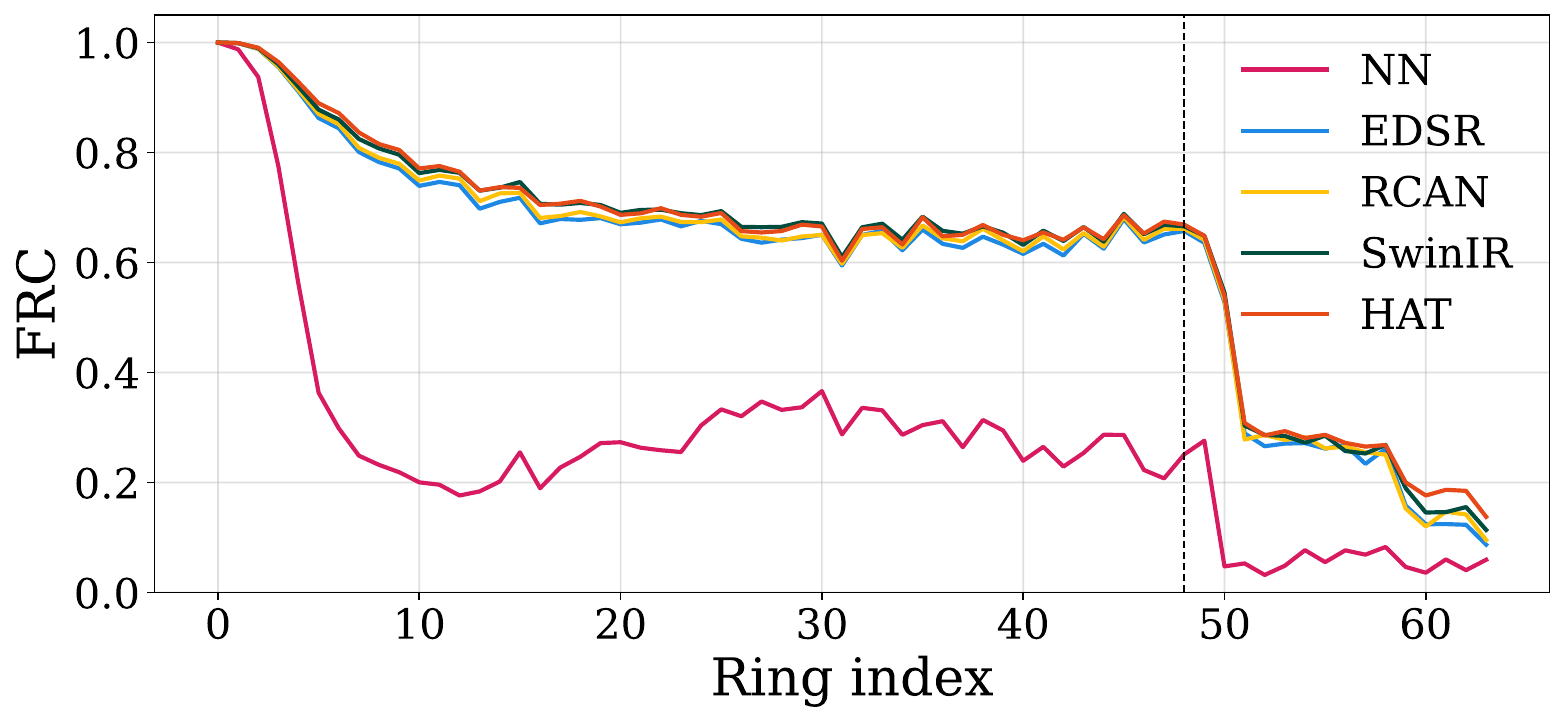} &
      \multirow{3}{*}[100pt]{\includegraphics[width=0.34\linewidth]{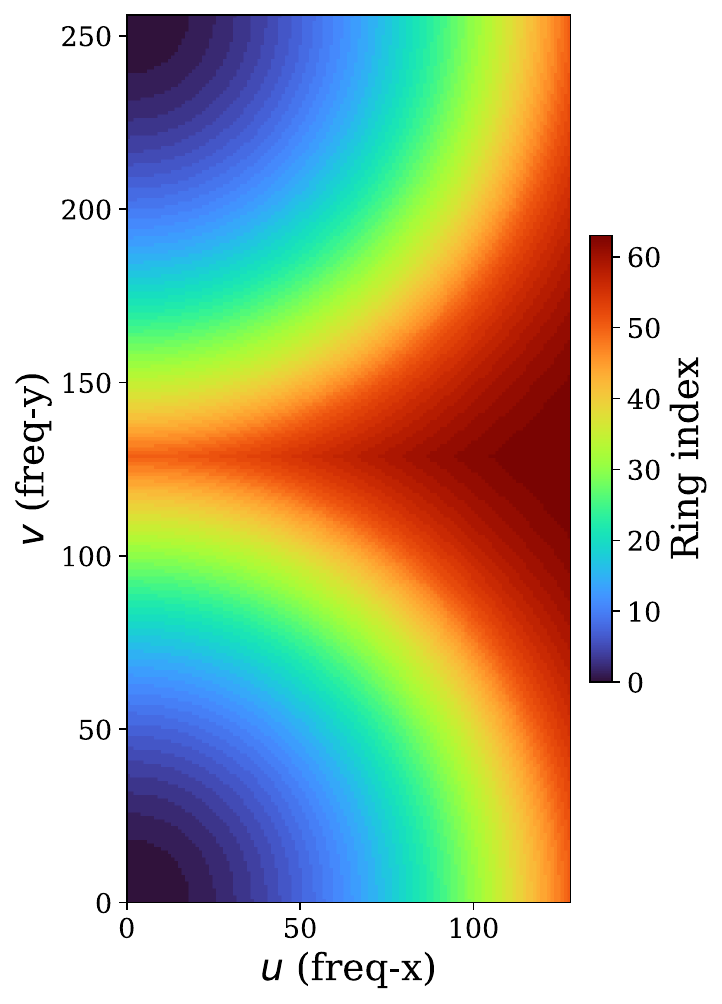}} \\

      \phantomsubcaption\label{fig:frc_graph_urban100}
      (b) Average FRC curve on Urban100 & \\

      \includegraphics[width=0.5\linewidth]{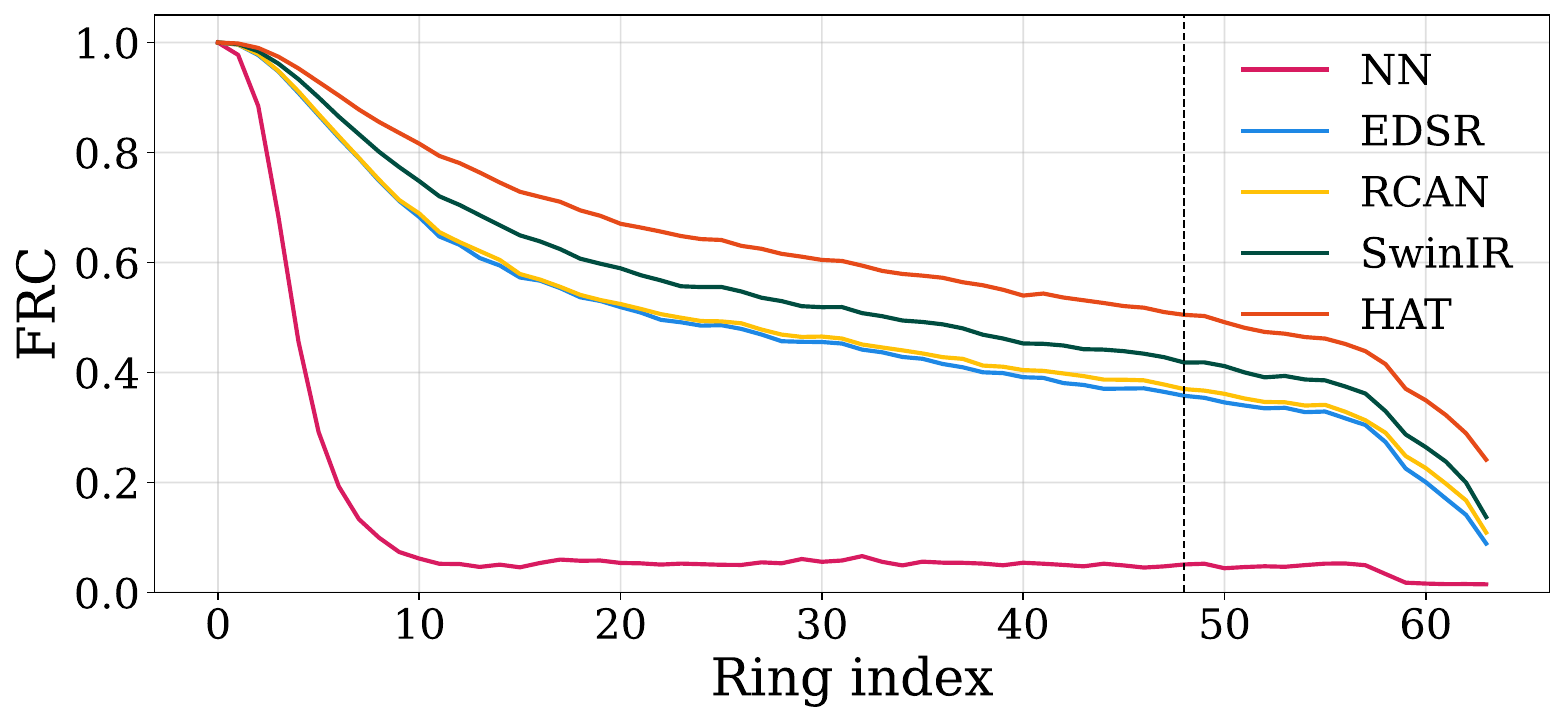} & 
      \phantomsubcaption\label{fig:frc_ring}
      \\
    \end{tabular}
    }
  \caption{%
    Graphs and frequency-ring indexing used in the FRC analysis, where NN denotes nearest-neighbor interpolation.
    \subref{fig:frc_graph_set5},\,\subref{fig:frc_graph_urban100}: Mean FRC curves on Set5 and Urban100.
    \subref{fig:frc_ring}: Visualization of the 64 quantized concentric rings used for spectral binning; colors indicate the ring indices.
    }
  \label{fig:frc_graphs_and_ring}
\end{figure*}

%% file: tables/frc_auc_results.tex
\begin{table}[h]
  \centering
  \setlength{\tabcolsep}{3pt}
  \scriptsize
    \caption{
    High-frequency reconstruction quality measured by \emph{FRC-AUC} on $\times$4 super-resolution benchmarks.
    For each backbone, the first row shows results using the baseline upsampler (SPC), and the second row uses the proposed FGA module.
    Higher scores indicate better preservation of high-frequency content.
    }
    \begin{tabular}{ll cccccc}
    \toprule
      \multirow{2}{*}[-2pt]{\textbf{Backbone}} &
      \multirow{2}{*}[-2pt]{\textbf{Upsampler}} &
      \multicolumn{1}{c}{\textbf{Set5}} &
      \multicolumn{1}{c}{\textbf{Set14}} &
      \multicolumn{1}{c}{\textbf{B100}} &
      \multicolumn{1}{c}{\textbf{U100}} &
      \multicolumn{1}{c}{\textbf{M109}} &
      \multicolumn{1}{c}{\textbf{D235}} \\
    \cmidrule(lr){3-8}
      & &
      \multicolumn{6}{c}{\scriptsize \textbf{\emph{FRC-AUC}}}  \\
    \midrule
    \midrule
    \multicolumn{1}{l}{} & SPC & 0.2648 & 0.2093 & 0.2126 & 0.2226 & 0.2998 & 0.1625 \\
    \multicolumn{1}{l}{\multirow{-2}{*}{EDSR\textsuperscript{$\mu$}}} & FGA \small{(ours)} & \textbf{0.2941} & \textbf{0.2347} & \textbf{0.2294} & \textbf{0.2804} & \textbf{0.3313} & \textbf{0.1758}  \\
    \midrule
    \multicolumn{1}{l}{} & SPC & 0.2857 & 0.2281 & 0.2279 & 0.2809 & 0.3221 & 0.1745 \\
    \multicolumn{1}{l}{\multirow{-2}{*}{EDSR}} & FGA \small{(ours)} & \textbf{0.2920} & \textbf{0.2358} & \textbf{0.2362} & \textbf{0.3010} & \textbf{0.3341} & \textbf{0.1771}  \\
    \midrule
    \multicolumn{1}{l}{} & SPC & 0.2740 & 0.2186 & 0.2199 & 0.2534 & 0.3216 & 0.1691 \\
    \multicolumn{1}{l}{\multirow{-2}{*}{RCAN\textsuperscript{$\mu$}}} & FGA \small{(ours)} & \textbf{0.3020} & \textbf{0.2373} & \textbf{0.2339} & \textbf{0.2975} & \textbf{0.3323} & \textbf{0.1771} \\
    \midrule
    \multicolumn{1}{l}{} & SPC & 0.2924 & 0.2367 & 0.2365 & 0.2962 & 0.3387 & 0.1788 \\
    \multicolumn{1}{l}{\multirow{-2}{*}{RCAN}} & FGA \small{(ours)} & \textbf{0.2947} & \textbf{0.2433} & \textbf{0.2389} & \textbf{0.3097} & \textbf{0.3424} & \textbf{0.1795}  \\
    \midrule
    \multicolumn{1}{l}{} & SPC & 0.2824 & 0.2289 & 0.2299 & 0.2799 & 0.3287 & 0.1735 \\
    \multicolumn{1}{l}{\multirow{-2}{*}{HAN\textsuperscript{$\mu$}}}  & FGA \small{(ours)} & \textbf{0.3028} & \textbf{0.2406} & \textbf{0.2378} & \textbf{0.3119} & \textbf{0.3487} & \textbf{0.1796} \\
    \midrule
    \multicolumn{1}{l}{} & SPC & 0.2211 & 0.1815 & 0.1983 & 0.1810 & 0.3224 & 0.1564 \\
    \multicolumn{1}{l}{\multirow{-2}{*}{HAN}}  & FGA \small{(ours)} & \textbf{0.2953} & \textbf{0.2397} & \textbf{0.2373} & \textbf{0.3070} & \textbf{0.3450} & \textbf{0.1776} \\
    \midrule
    \multicolumn{1}{l}{} & SPC & 0.2657 & 0.2155 & 0.2144 & 0.2330 & 0.3142 & 0.1670 \\
    \multicolumn{1}{l}{\multirow{-2}{*}{NLSN\textsuperscript{$\mu$}}} & FGA \small{(ours)} & \textbf{0.2963} & \textbf{0.2439} & \textbf{0.2366} & \textbf{0.3017} & \textbf{0.3439} & \textbf{0.1799} \\
    \midrule
    \multicolumn{1}{l}{} & SPC & 0.2851 & 0.2354 & 0.2357 & 0.3019 & 0.3380 & 0.1769 \\
    \multicolumn{1}{l}{\multirow{-2}{*}{NLSN}} & FGA \small{(ours)} & \textbf{0.2999} & \textbf{0.2440} & \textbf{0.2415} & \textbf{0.3178} & \textbf{0.3502} & \textbf{0.1820} \\
    \midrule
    \multicolumn{1}{l}{} & SPC & 0.2655 & 0.1902 & 0.2209 & 0.2446 & 0.3342 & 0.1641 \\
    \multicolumn{1}{l}{\multirow{-2}{*}{SwinIR\textsuperscript{$\mu$}}} & FGA \small{(ours)} & \textbf{0.2965} & \textbf{0.2286} & \textbf{0.2392} & \textbf{0.3080} & \textbf{0.3548} & \textbf{0.1740} \\
    \midrule
    \multicolumn{1}{l}{} & SPC & 0.2955 & 0.2238 & 0.2470 & 0.3371 & 0.3616 & 0.1805 \\
    \multicolumn{1}{l}{\multirow{-2}{*}{SwinIR}} & FGA \small{(ours)} & \textbf{0.3028} & \textbf{0.2336} & \textbf{0.2492} & \textbf{0.3477} & \textbf{0.3715} & \textbf{0.1815} \\
    \bottomrule
    \end{tabular}%
  \label{tab:frc_auc_results}
\end{table}

%% file: figs/hf_vis_results.tex
\begin{figure*}[!t]
    \renewcommand{\arraystretch}{1.1}
    \renewcommand{\tabcolsep}{0.9mm}
    \resizebox{\textwidth}{!}{%
        \begin{tabular}{ccccccccccc}
            \phantomsubcaption\label{fig:hf-vis-lw-scale}
            & &  EDSR &  RCAN &  HAN &  NLSN &  SwinIR &&&  HR (top) / Cropped HR (bottom) \\
            \multirow{2}{*}[35pt]{\rotatebox{90}{\textbf{(a) Lightweight scale}}}
            & \multirow{1}{*}[35pt]{\rotatebox{90}{ SPC}}
            & \includegraphics[height=0.15\linewidth, width=0.15\linewidth]{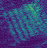}
            & \includegraphics[height=0.15\linewidth, width=0.15\linewidth]{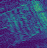}
            & \includegraphics[height=0.15\linewidth, width=0.15\linewidth]{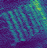}
            & \includegraphics[height=0.15\linewidth, width=0.15\linewidth]{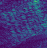}
            & \includegraphics[height=0.15\linewidth, width=0.15\linewidth]{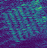} 
            & \multicolumn{1}{c|}{}
            & \multicolumn{1}{|c}{}
            & \multirow{2}{*}[65pt]{\centering\includegraphics[height=0.31\linewidth, width=0.31\linewidth]{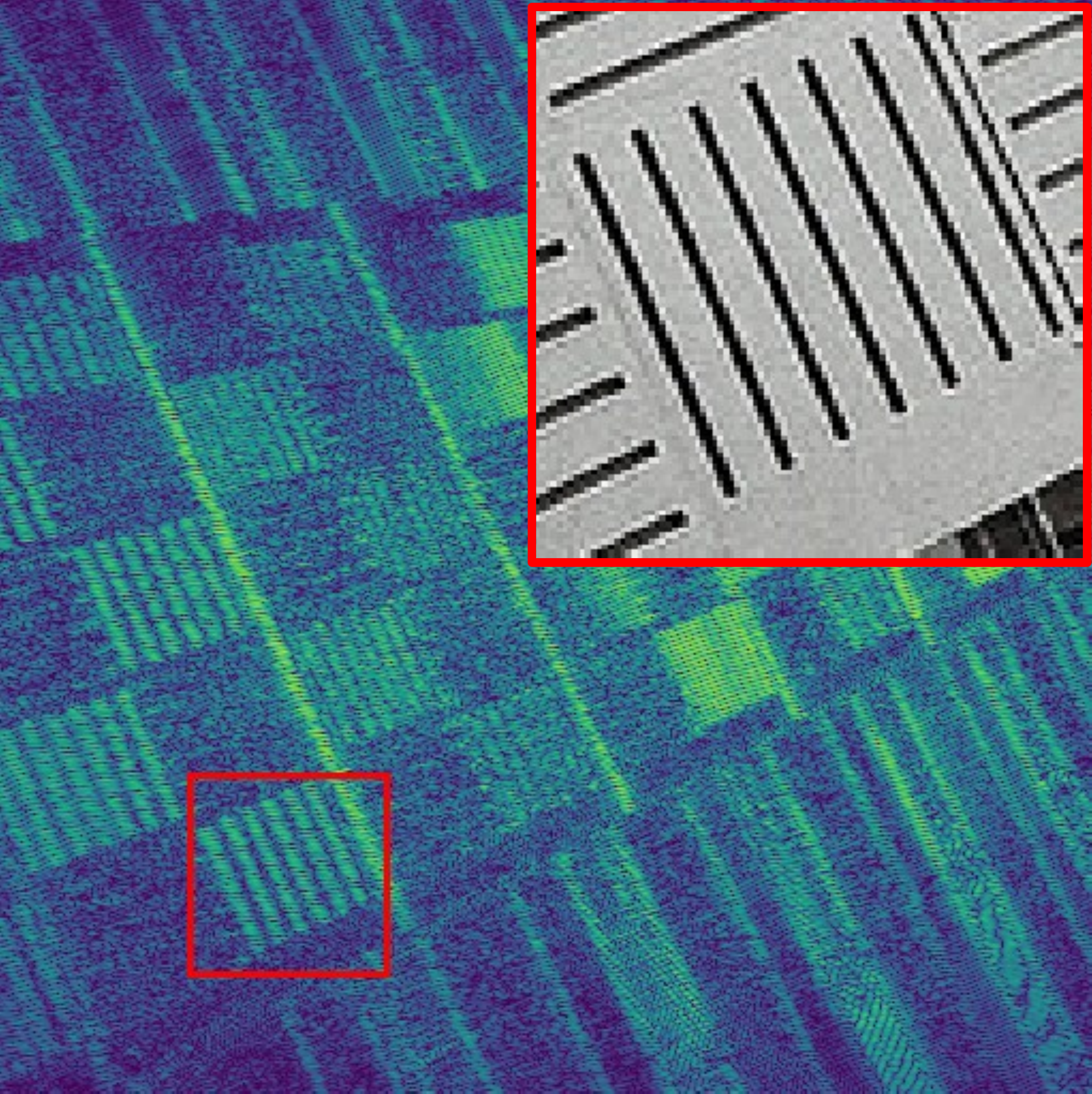}} \\
            
            & \multirow{1}{*}[52pt]{\rotatebox{90}{ FGA (ours)}}
            & \includegraphics[height=0.15\linewidth, width=0.15\linewidth]{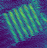}
            & \includegraphics[height=0.15\linewidth, width=0.15\linewidth]{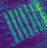}
            & \includegraphics[height=0.15\linewidth, width=0.15\linewidth]{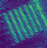}
            & \includegraphics[height=0.15\linewidth, width=0.15\linewidth]{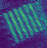}
            & \includegraphics[height=0.15\linewidth, width=0.15\linewidth]{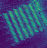} 
            & \multicolumn{1}{c|}{}
            & \multicolumn{1}{|c}{}
            & \\  [15pt]

            \phantomsubcaption\label{fig:hf-vis-origin-scale}
            \multirow{2}{*}[22pt]{\rotatebox{90}{\textbf{(b) Original scale}}} 
            & \multirow{1}{*}[35pt]{\rotatebox{90}{ SPC}}
            & \includegraphics[height=0.15\linewidth, width=0.15\linewidth]{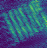}
            & \includegraphics[height=0.15\linewidth, width=0.15\linewidth]{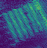}
            & \includegraphics[height=0.15\linewidth, width=0.15\linewidth]{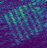}
            & \includegraphics[height=0.15\linewidth, width=0.15\linewidth]{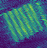}
            & \includegraphics[height=0.15\linewidth, width=0.15\linewidth]{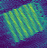} 
            & \multicolumn{1}{c|}{} 
            & \multicolumn{1}{|c}{}
            & \multirow{2}{*}[65pt]{\centering\includegraphics[height=0.31\linewidth, width=0.31\linewidth]{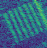}} \\
            
            & \multirow{1}{*}[52pt]{\rotatebox{90}{ FGA (ours)}}
            & \includegraphics[height=0.15\linewidth, width=0.15\linewidth]{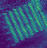}
            & \includegraphics[height=0.15\linewidth, width=0.15\linewidth]{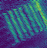}
            & \includegraphics[height=0.15\linewidth, width=0.15\linewidth]{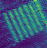}
            & \includegraphics[height=0.15\linewidth, width=0.15\linewidth]{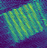}
            & \includegraphics[height=0.15\linewidth, width=0.15\linewidth]{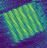} 
            & \multicolumn{1}{c|}{} 
            & \multicolumn{1}{|c}{}
            & & 
        \end{tabular}
        }
    \caption{
    High-frequency magnitude maps visualized from the top quartile of the Fourier spectrum (rings 48–63) for ``img\_092'' in Urban100~\cite{Urban100}.
    Columns show outputs from different backbones, followed by the ground-truth high-resolution image and a zoomed-in view of the red-boxed region.
    Panel \subref{fig:hf-vis-lw-scale} shows results for lightweight models, while panel \subref{fig:hf-vis-origin-scale} presents original-scale models as defined in the respective papers.
    }
    \label{fig:hf_vis_results}

\end{figure*} 

%% file: tables/ablation_study.tex
\begin{table}[h]
    \centering
    \scriptsize
    \setlength{\tabcolsep}{2pt}
    \caption{
    Ablation study of the proposed FGA upsampler using the EDSR$^\mu$ backbone.
    We report PSNR, SSIM, and high-frequency FRC-AUC on Set5 and Urban100.
    }
    \begin{tabular}{cccccccccccc}
        \toprule
        \multicolumn{4}{c}{\textbf{Module}} & \multicolumn{2}{c}{\textbf{Loss}} & \multicolumn{3}{c}{\textbf{Set5}} & \multicolumn{3}{c}{\textbf{Urban100}} \\
        \cmidrule(lr){1-4}\cmidrule(lr){5-6}\cmidrule(lr){7-9}\cmidrule(lr){10-12}
        \textbf{Conv} & \textbf{MLP} & \textbf{FF} & \textbf{CAL} & \textbf{L1} & \textbf{FL1} & \textbf{PSNR} & \textbf{SSIM} & \textbf{FRC} & \textbf{PSNR} & \textbf{SSIM} & \textbf{FRC} \\
        \midrule
        \checkmark & \checkmark & \checkmark & \checkmark & & \checkmark & \textbf{32.30} & \textbf{0.8966} & \textbf{0.2934} & \textbf{26.19} & \textbf{0.7867} & \textbf{0.2743} \\
        \midrule
        \checkmark & \checkmark & \checkmark & \checkmark & \checkmark & & 32.18 & 0.8956 & 0.2792 & 26.09 & 0.7864 & 0.2616 \\
        \midrule
        \checkmark & \checkmark & \checkmark &  & \checkmark & & 32.16 & 0.8951 & 0.2806 & 26.04 & 0.7843 & 0.2527 \\
        \midrule
        \checkmark & \checkmark &  &  & \checkmark & & 32.11 & 0.8948 & 0.2673 & 26.00 & 0.7834 & 0.2359 \\
        \midrule
        \checkmark &  &  &  & \checkmark & & 32.12 & 0.8947 & 0.2555 & 26.00 & 0.7833 & 0.2111 \\
        \bottomrule
    \end{tabular}
    \label{tab:ablation_study}
\end{table}

%% file: sec/10_conclusion.tex
\section{Conclusion}
\label{sec:conclusion}

We proposed a novel frequency-guided attention (FGA) upsampler for image super-resolution, designed to suppress aliasing and enhance high-frequency detail through explicit spatial and spectral conditioning. 
By injecting Fourier-encoded sub-pixel coordinates and employing correlation-based attention, FGA enables each upsampled pixel to learn position-aware, frequency-sensitive representations.

Extensive experiments on standard benchmarks demonstrate that FGA consistently outperforms conventional SPC across a wide range of backbones and model scales, yielding improvements in PSNR, SSIM, and frequency-domain similarity.
We further validated its generalization across lightweight and full-capacity networks and provided spectral visualizations that explain FGA’s role in mitigating artifacts such as checkerboard patterns and spectral folding.
Ablation studies highlighted the complementary contributions of the Fourier features, attention layer, and frequency-domain loss.

\paragraph{Limitations and Future Work}

While FGA significantly improves reconstruction quality, it incurs a notable increase in computational overhead relative to SPC. As shown in Tab.~\ref{tab:comp_cost}, FLOPs and memory usage increase by approximately 2-4$\times$ depending on the input resolution.
Nevertheless, we find that allocating capacity to the upsampling module leads to greater performance gains than simply expanding the backbone.

Although the computational overhead introduced by FGA is relatively modest, accounting for only about 5\% of the total FLOPs in large-scale SR networks such as EDSR, reducing the cost of Fourier-guided spatial modulation remains an important direction for future work.
One promising approach is to leverage hardware accelerators such as the Transformer Engine~\cite{TE}, which can help reduce latency and improve throughput, particularly for real-time or resource-constrained applications.
\input{tables/com_results}

In contrast, the memory consumption of FGA is significantly higher than that of conventional upsamplers like SPC, as shown in Tab.~\ref{tab:comp_cost}.
While modern GPUs for deep learning typically offer sufficient memory to accommodate this increase, the elevated memory usage may still be a bottleneck in resource-limited environments. 
Therefore, reducing memory overhead through lightweight design, attention pruning, or optimized data flow may be a valuable direction for future improvements.

\input{figs/feat_results_swinir}

%% file: tables/com_results.tex
\begin{table}[h]
  \centering
  \scriptsize
  \setlength{\tabcolsep}{4pt}
  \caption{Computational cost of \textbf{backbone} and \textbf{upsampler}
           at four input resolutions.}
  \label{tab:comp_cost}
  \begin{tabular}{l rrrr @{\hspace{8pt}} rrrrr}
    \toprule
        & \multicolumn{4}{c}{\textbf{GFLOPs}}
        &  & \multicolumn{4}{c}{\textbf{Memory (MB)}} \\
    \cmidrule(lr){2-5}\cmidrule(lr){7-10}
    \textbf{Resolution}
        & {$64^{2}$} & {$128^{2}$} & {$256^{2}$} & {$512^{2}$}
        &  & {$64^{2}$} & {$128^{2}$} & {$256^{2}$} & {$512^{2}$} \\[0.15em]
    \cmidrule(lr){1-10}
    \multicolumn{10}{l}{\textbf{Backbone}} \\[-0.2em]
    \cmidrule(lr){1-10}
    EDSR   & 157 & 628 & 2,513 & 10,052 && 173 & 245 & 469   & 1,432 \\
    RCAN   &  62 & 248 &   993 &  3,972 &&  64 &  83 & 156   &   446 \\
    HAN    &  64 & 256 & 1,026 &  4,102 && 118 & 253 & 806   & 3,016 \\
    NLSN   & 171 & 682 & 2,729 & 10,917 && 350 & 911 & 3,168 & 12,180 \\
    SwinIR &  49 & 176 &   681 &  2,701 && 118 & 245 & 739   & 2,708 \\
    \addlinespace[0.6ex]
    \multicolumn{10}{l}{\textbf{Upsampler}} \\[-0.2em]
    \cmidrule(lr){1-10}
    SPC    &  3 & 13 &  53 & 210 &&  44 & 148 & 561 & 2,217 \\
    FGA    &  8 & 30 & 120 & 482 && 155 & 576 & 2,259 & 8,991 \\
    \bottomrule
  \end{tabular}
\end{table}

%% file: figs/feat_results_swinir.tex
\begin{figure*}[!t]
    \renewcommand{\arraystretch}{1.1}
    \renewcommand{\tabcolsep}{0.9mm}
    \centering
    \resizebox{\textwidth}{!}{%
        \begin{tabular}{ccccccccccc}
            \phantomsubcaption\label{fig:feat_spec_swinir}
            & &  Interp+Conv &  Deconv &  PixelShuffle &  Ours (FGA) &&&  GT &  HR (bottom) / LR (top) \\
            \multirow{2}{*}[15pt]{\rotatebox{90}{\textbf{(a) Spectrum}}}
            & \multirow{1}{*}[38pt]{\rotatebox{90}{ Pre-}}
            & \includegraphics[height=0.15\linewidth, width=0.15\linewidth]{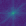}
            & \includegraphics[height=0.15\linewidth, width=0.15\linewidth]{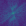}
            & \includegraphics[height=0.15\linewidth, width=0.15\linewidth]{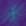}
            & \includegraphics[height=0.15\linewidth, width=0.15\linewidth]{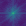}
            & \multicolumn{1}{c|}{}
            & \multicolumn{1}{|c}{}
            & \includegraphics[height=0.15\linewidth, width=0.15\linewidth]{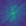} 
            & \multirow{2}{*}[65pt]{\centering\includegraphics[height=0.31\linewidth, keepaspectratio=true]{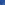}} \\
            
            & \multirow{1}{*}[40pt]{\rotatebox{90}{ Post-}}
            & \includegraphics[height=0.15\linewidth, width=0.15\linewidth]{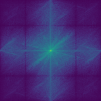} 
            & \includegraphics[height=0.15\linewidth, width=0.15\linewidth]{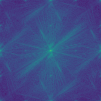}
            & \includegraphics[height=0.15\linewidth, width=0.15\linewidth]{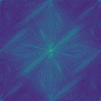}
            & \includegraphics[height=0.15\linewidth, width=0.15\linewidth]{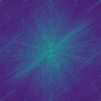}
            & \multicolumn{1}{c|}{} 
            & \multicolumn{1}{|c}{}
            & \includegraphics[height=0.15\linewidth, width=0.15\linewidth]{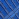} 
            & \\  [15pt]

            \phantomsubcaption\label{fig:feat_map_swinir}
            \multirow{2}{*}[20pt]{\rotatebox{90}{\textbf{(b) Feature map}}} 
            & \multirow{1}{*}[38pt]{\rotatebox{90}{ Pre-}}
            & \includegraphics[height=0.15\linewidth, width=0.15\linewidth]{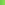}
            & \includegraphics[height=0.15\linewidth, width=0.15\linewidth]{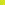}
            & \includegraphics[height=0.15\linewidth, width=0.15\linewidth]{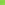}
            & \includegraphics[height=0.15\linewidth, width=0.15\linewidth]{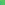}
            & \multicolumn{1}{c|}{} 
            & \multicolumn{1}{|c}{}
            & \includegraphics[height=0.15\linewidth, width=0.15\linewidth]{imgs/features/SwinIR-M/feat/cropped_img_012_lr.pdf} 
            & \multirow{2}{*}[65pt]{\centering\includegraphics[height=0.31\linewidth, keepaspectratio=true]{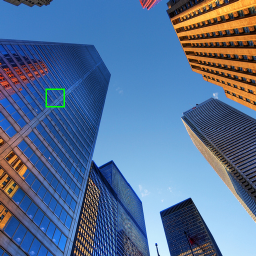}} \\
            
            & \multirow{1}{*}[40pt]{\rotatebox{90}{ Post-}}
            & \includegraphics[height=0.15\linewidth, width=0.15\linewidth]{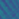}
            & \includegraphics[height=0.15\linewidth, width=0.15\linewidth]{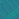}
            & \includegraphics[height=0.15\linewidth, width=0.15\linewidth]{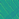}
            & \includegraphics[height=0.15\linewidth, width=0.15\linewidth]{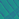}
            & \multicolumn{1}{c|}{} 
            & \multicolumn{1}{|c}{}
            & \includegraphics[height=0.15\linewidth, width=0.15\linewidth]{imgs/features/SwinIR-M/feat/cropped_img_012_gt.pdf} 
            & & 
        \end{tabular}
        }
    \caption{
    Feature and frequency spectrum visualization of different upsampling methods using the SwinIR\textsuperscript{$\mu$} backbone on ``img\_012'' from Urban100~\cite{Urban100}. 
    The input is center-cropped to $128 \times 128$ pixels and processed with $\times4$ SR inference. 
    The top two rows display feature maps before and after upsampling, while the bottom two rows show their corresponding Fourier spectra.
    }
\label{fig:feat_results_swinir}

\end{figure*} 

%% file: sec/12_appendix.tex
\input{figs/han_results}

\section{Effect of Window Size}\label{app:window_size}
\input{tables/winsize_exp}
We evaluate various local attention window sizes used in the cross-resolution attention module to identify the optimal configuration for balancing performance and efficiency.
Using the EDSR$^\mu$ backbone, we test multiple pre- and post-upsampling window pairs on Set5 and Urban100, and also profile their FLOPs and memory usage on $512^2$ resolution inputs (Tab.~\ref{tab:window_size_comparison}).

Among all tested settings, the $(5 \times 5,~4 \times 4)$ overlapping-window configuration achieves the best trade-off, offering improved PSNR/SSIM with only a moderate increase in computational cost.
This configuration is thus adopted as the default throughout the main experiments.

\section{FRC-AUC Result Analysis for HAN}\label{app:han}

While our proposed FGA consistently improves FRC-AUC across most backbones, we observe an interesting anomaly with the HAN backbone. 
The original HAN model (initialized from publicly released checkpoints) occasionally produces halo or ringing artifacts around edges (Fig.~\ref{fig:han_results}). 
In the frequency domain, these appear as off-diagonal or radial ridges, while in the spatial domain they manifest as visible halos in image crops. 
Such artifacts introduce excess high-frequency energy that deviates from the ground truth, thereby lowering FRC-AUC even when PSNR and SSIM remain high.

By contrast, FGA reduces excess high-frequency energy and brings the spectrum closer to GT, while the from-scratch HAN$^\mu$ is notably more resistant to such artifacts.
Integrating FGA into either variant reduces excess high-frequency energy and aligns the spectrum more closely with the ground truth, demonstrating that frequency-domain correlation provides a complementary perspective to conventional pixel-space metrics.

%% file: figs/han_results.tex
\begin{figure*}[!t]
\centering
\setlength{\tabcolsep}{0.7mm}
\setcounter{subfigure}{0}
\captionsetup[subfloat]{font=small}
\begin{adjustbox}{max totalsize={\textwidth}{0.90\textheight}}

\subfloat[HAN results of img\_025]{%
\begin{adjustbox}{max size={\textwidth}{0.44\textheight}}
\begin{tabular}{cccccccc}
    & &  HAN$^\mu$ &  HAN &&&  GT \\
    \multirow{2}{*}[15pt]{\normalsize \rotatebox{90}{ \textbf{Spectrum}}}
    & \multirow{1}{*}[29pt]{\rotatebox{90}{ SPC}}
    & \includegraphics[height=0.15\linewidth, width=0.15\linewidth]{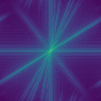}
    & \includegraphics[height=0.15\linewidth, width=0.15\linewidth]{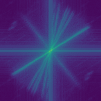}
    & \multicolumn{1}{c|}{}
    & \multicolumn{1}{|c}{}
    & \multirow{2}{*}[65pt]{\centering\includegraphics[height=0.31\linewidth, keepaspectratio=true]{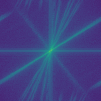}} \\
    &
    \multirow{1}{*}[34pt]{\rotatebox{90}{ FGA}}
    & \includegraphics[height=0.15\linewidth, width=0.15\linewidth]{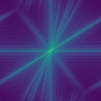} 
    & \includegraphics[height=0.15\linewidth, width=0.15\linewidth]{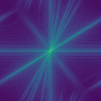}
    & \multicolumn{1}{c|}{} 
    & \multicolumn{1}{|c}{}
    & \\
    \multirow{2}{*}[5pt]{\normalsize \rotatebox{90}{ \textbf{Image}}}
    & \multirow{1}{*}[30pt]{\rotatebox{90}{ SPC}}
    & \includegraphics[height=0.15\linewidth, width=0.15\linewidth]{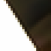}
    & \includegraphics[height=0.15\linewidth, width=0.15\linewidth]{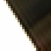}
    & \multicolumn{1}{c|}{}
    & \multicolumn{1}{|c}{}
    & \multirow{2}{*}[65pt]{\centering\includegraphics[height=0.31\linewidth, keepaspectratio=true]{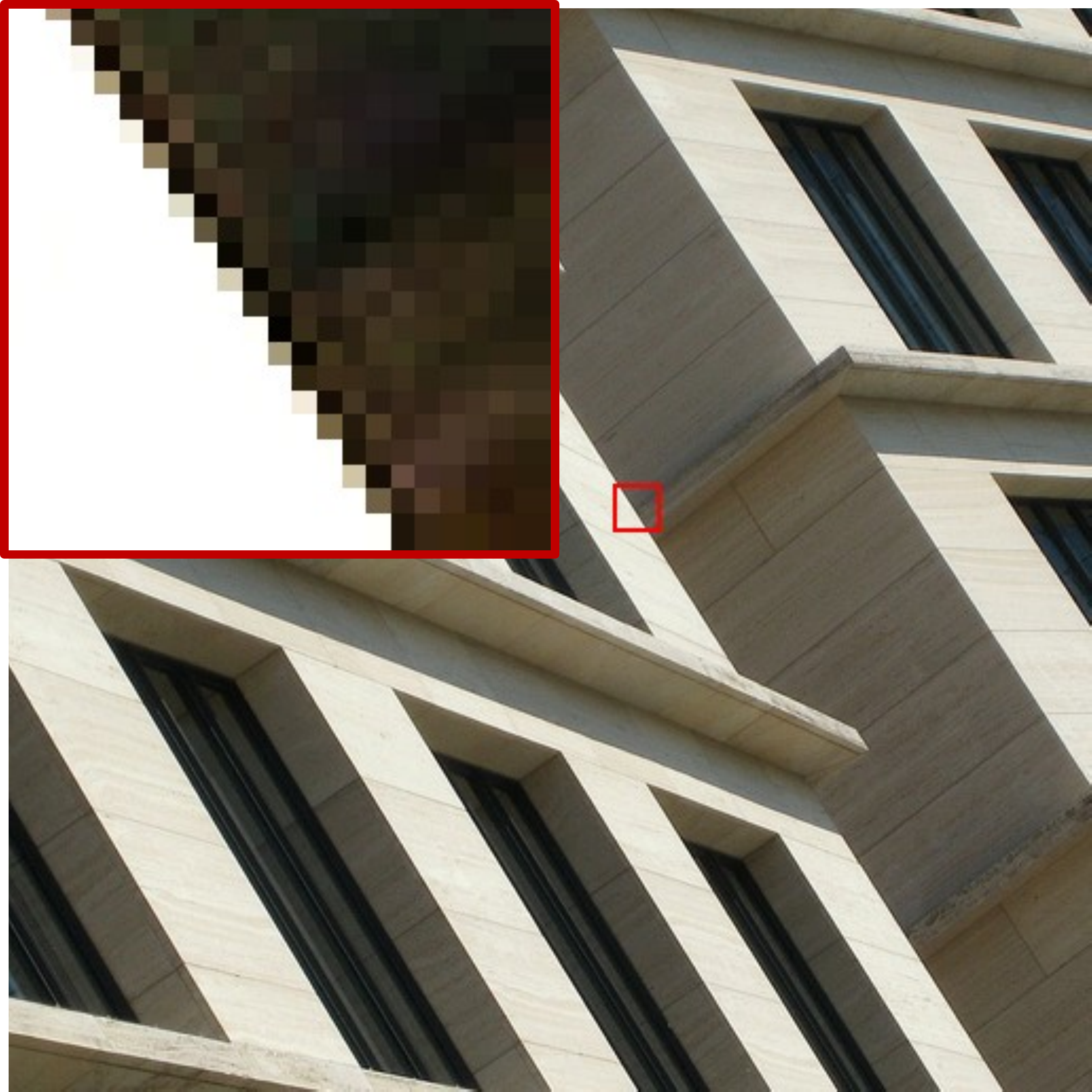}} \\
    &
    \multirow{1}{*}[37pt]{\rotatebox{90}{ FGA}}
    & \includegraphics[height=0.15\linewidth, width=0.15\linewidth]{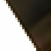}
    & \includegraphics[height=0.15\linewidth, width=0.15\linewidth]{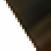}
    & \multicolumn{1}{c|}{}
    & \multicolumn{1}{|c}{}
    & &
\end{tabular}
\end{adjustbox}
}\vspace{4pt}
\subfloat[HAN restults of img\_059]{%
\begin{adjustbox}{max size={\textwidth}{0.44\textheight}}
\begin{tabular}{cccccccc}
    & &  HAN$^\mu$ &  HAN &&&  GT \\
    \multirow{2}{*}[15pt]{\normalsize \rotatebox{90}{ \textbf{Spectrum}}}
    & \multirow{1}{*}[29pt]{\rotatebox{90}{ SPC}}
    & \includegraphics[height=0.15\linewidth, width=0.15\linewidth]{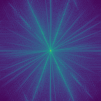}
    & \includegraphics[height=0.15\linewidth, width=0.15\linewidth]{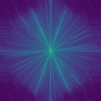}
    & \multicolumn{1}{c|}{}
    & \multicolumn{1}{|c}{}
    & \multirow{2}{*}[65pt]{\centering\includegraphics[height=0.31\linewidth, keepaspectratio=true]{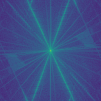}} \\
    &
    \multirow{1}{*}[34pt]{\rotatebox{90}{ FGA}}
    & \includegraphics[height=0.15\linewidth, width=0.15\linewidth]{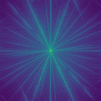} 
    & \includegraphics[height=0.15\linewidth, width=0.15\linewidth]{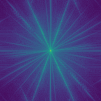}
    & \multicolumn{1}{c|}{} 
    & \multicolumn{1}{|c}{}
    & \\
    \multirow{2}{*}[5pt]{\normalsize \rotatebox{90}{ \textbf{Image}}} 
    & \multirow{1}{*}[30pt]{\rotatebox{90}{ SPC}}
    & \includegraphics[height=0.15\linewidth, width=0.15\linewidth]{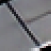}
    & \includegraphics[height=0.15\linewidth, width=0.15\linewidth]{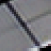}
    & \multicolumn{1}{c|}{} 
    & \multicolumn{1}{|c}{}
    & \multirow{2}{*}[65pt]{\centering\includegraphics[height=0.31\linewidth, keepaspectratio=true]{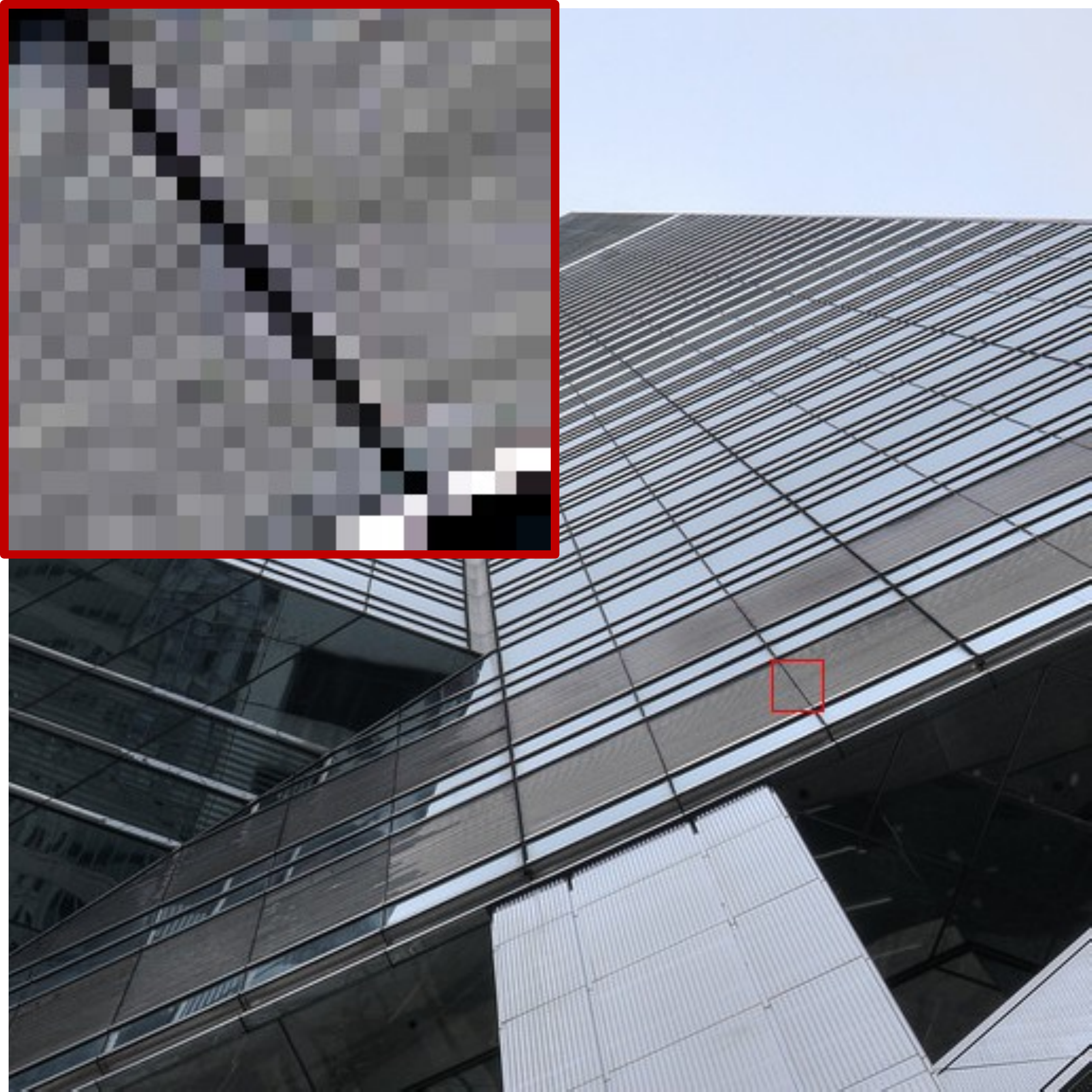}} \\
    &
    \multirow{1}{*}[37pt]{\rotatebox{90}{ FGA}}
    & \includegraphics[height=0.15\linewidth, width=0.15\linewidth]{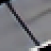}
    & \includegraphics[height=0.15\linewidth, width=0.15\linewidth]{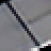}
    & \multicolumn{1}{c|}{} 
    & \multicolumn{1}{|c}{}
    & &
\end{tabular}
\end{adjustbox}
}
\end{adjustbox}
\caption{
    Two Urban100 cases comparing HAN and HAN$^\mu$ under SPC and FGA. 
    Panels (a)–(b) correspond to ``img\_025'' and ``img\_059'', respectively. 
    For each case, the \textbf{top} row shows the Fourier magnitude spectrum, and the \textbf{bottom} row shows spatial crops (ground truth on the right; red boxes indicate zoomed regions). 
    For the spatial crops, contrast has been enhanced to make halo artifacts more visible.
}
\label{fig:han_results}
\end{figure*}

%% file: tables/winsize_exp.tex
\newcommand{\cell}[1]{\makebox[8.5mm][c]{#1}}
\begin{table}[h]
  \centering
  \footnotesize
  \caption{Impact of local attention window size on image quality and computational efficiency. 
  The first two columns indicate the window sizes applied before and after upsampling, respectively. 
  Each configuration is evaluated on Set5 and Urban100 in terms of PSNR and SSIM, along with FLOPs and peak memory usage measured at $512^2$ input resolution.
  }
  \label{tab:window_size_comparison}
  \setlength{\tabcolsep}{3pt}
  \renewcommand{\arraystretch}{1.07}

  \begin{tabular}{cc*{6}{c}}
    \toprule
    \multicolumn{2}{c}{\textbf{Window Size}} &
      \multicolumn{2}{c}{\textbf{Set5}} &
      \multicolumn{2}{c}{\textbf{Urban100}} &
      \multicolumn{2}{c}{$\boldsymbol{512^{2}}$ \textbf{res}} \\
    \cmidrule(lr){1-2}\cmidrule(lr){3-4}\cmidrule(lr){5-6}\cmidrule(lr){7-8}
    \makecell{Pre\\up} & \makecell{Post\\up} &
      \cell{PSNR} & \cell{SSIM} & \cell{PSNR} & \cell{SSIM} &
      \cell{GFLOPs} & \cell{Mem.} \\
    \midrule
    3 × 3 & 4 × 4 & \cell{32.11} & \cell{0.8954} & \cell{26.09} & \cell{0.7860} & \cell{90.73}  & \cell{6,592} \\
    \textbf{5 × 5} & \textbf{4 × 4} & \cell{\textbf{32.19}} & \cell{\textbf{0.8955}} & \cell{\textbf{26.10}} & \cell{\textbf{0.7863}} & \cell{\textbf{99.32}} & \cell{\textbf{8,896}} \\
    7 × 7 & 4 × 4 & \cell{32.13} & \cell{0.8951} & \cell{26.09} & \cell{0.7861} & \cell{112.21} & \cell{14,768} \\
    9 × 9 & 4 × 4 & \cell{32.19} & \cell{0.8959} & \cell{26.09} & \cell{0.7861} & \cell{129.39} & \cell{22,960} \\
    \midrule
    2 × 2 & 8 × 8 & \cell{32.15} & \cell{0.8953} & \cell{26.09} & \cell{0.7861} & \cell{88.05}  & \cell{6,280} \\
    4 × 4 & 8 × 8 & \cell{32.11} & \cell{0.8950} & \cell{26.09} & \cell{0.7860} & \cell{94.49}  & \cell{6,216} \\
    8 × 8 & 8 × 8 & \cell{32.12} & \cell{0.8956} & \cell{26.09} & \cell{0.7859} & \cell{120.26} & \cell{8,368} \\
    \midrule
   16 × 16 & 8 × 8  & \cell{32.20} & \cell{0.8958} & \cell{26.09} & \cell{0.7862} & \cell{223.34} & \cell{19,632} \\
    4 × 4  &16 × 16 & \cell{32.15} & \cell{0.8955} & \cell{26.10} & \cell{0.7861} & \cell{94.49}  & \cell{6,280} \\
    8 × 8  &32 × 32 & \cell{32.14} & \cell{0.8955} & \cell{26.09} & \cell{0.7860} & \cell{120.26} & \cell{6,384} \\
    \bottomrule
  \end{tabular}
\end{table}

%% file: main.bbl
\begin{thebibliography}{59}
\providecommand{\natexlab}[1]{#1}
\providecommand{\url}[1]{\texttt{#1}}
\expandafter\ifx\csname urlstyle\endcsname\relax
  \providecommand{\doi}[1]{doi: #1}\else
  \providecommand{\doi}{doi: \begingroup \urlstyle{rm}\Url}\fi

\bibitem[Agnihotri et~al.(2024)Agnihotri, Grabinski, and Keuper]{LCTC}
Shashank Agnihotri, Julia Grabinski, and Margret Keuper.
\newblock Improving feature stability during upsampling--spectral artifacts and the importance of spatial context.
\newblock In \emph{European Conference on Computer Vision}, pages 357--376. Springer, 2024.

\bibitem[Agustsson and Timofte(2017)]{DIV2K}
Eirikur Agustsson and Radu Timofte.
\newblock Ntire 2017 challenge on single image super-resolution: Dataset and study.
\newblock In \emph{Proceedings of the IEEE conference on computer vision and pattern recognition workshops}, pages 126--135, 2017.

\bibitem[Bevilacqua et~al.(2012)]{Set5}
Marco Bevilacqua et~al.
\newblock Low-complexity single-image super-resolution based on nonnegative neighbor embedding.
\newblock pages 135--1, 2012.

\bibitem[Chen et~al.(2021{\natexlab{a}})Chen, Wang, Guo, Xu, Deng, Liu, Ma, Xu, Xu, and Gao]{IPT}
Hanting Chen, Yunhe Wang, Tianyu Guo, Chang Xu, Yiping Deng, Zhenhua Liu, Siwei Ma, Chunjing Xu, Chao Xu, and Wen Gao.
\newblock Pre-trained image processing transformer.
\newblock In \emph{Proceedings of the IEEE/CVF Conference on Computer Vision and Pattern Recognition (CVPR)}, pages 12299--12310, 2021{\natexlab{a}}.

\bibitem[Chen et~al.(2023{\natexlab{a}})Chen, Wang, Zhou, Qiao, and Dong]{HAT}
Xiangyu Chen, Xintao Wang, Jiantao Zhou, Yu Qiao, and Chao Dong.
\newblock Activating more pixels in image super-resolution transformer.
\newblock In \emph{Proceedings of the IEEE/CVF Conference on Computer Vision and Pattern Recognition (CVPR)}, pages 22367--22377, 2023{\natexlab{a}}.

\bibitem[Chen et~al.(2021{\natexlab{b}})Chen, Liu, and Wang]{LIIF}
Yinbo Chen, Sifei Liu, and Xiaolong Wang.
\newblock Learning continuous image representation with local implicit image function.
\newblock In \emph{Proceedings of the IEEE/CVF Conference on Computer Vision and Pattern Recognition (CVPR)}, pages 8628--8638, 2021{\natexlab{b}}.

\bibitem[Chen(2019)]{INR_2}
Zhiqin Chen.
\newblock Im-net: Learning implicit fields for generative shape modeling.
\newblock 2019.

\bibitem[Chen et~al.(2022)Chen, Zhang, Gu, zhang, Kong, and Yuan]{CAT}
Zheng Chen, Yulun Zhang, Jinjin Gu, yongbing zhang, Linghe Kong, and Xin Yuan.
\newblock Cross aggregation transformer for image restoration.
\newblock In \emph{Advances in Neural Information Processing Systems}, pages 25478--25490. Curran Associates, Inc., 2022.

\bibitem[Chen et~al.(2023{\natexlab{b}})Chen, Zhang, Gu, Kong, Yang, and Yu]{DAT}
Zheng Chen, Yulun Zhang, Jinjin Gu, Linghe Kong, Xiaokang Yang, and Fisher Yu.
\newblock Dual aggregation transformer for image super-resolution.
\newblock In \emph{Proceedings of the IEEE/CVF international conference on computer vision}, pages 12312--12321, 2023{\natexlab{b}}.

\bibitem[Cho et~al.(2014)Cho, van Merrienboer, Gulcehre, Bahdanau, Bougares, Schwenk, and Bengio]{Attention}
Kyunghyun Cho, Bart van Merrienboer, Caglar Gulcehre, Dzmitry Bahdanau, Fethi Bougares, Holger Schwenk, and Yoshua Bengio.
\newblock Learning phrase representations using rnn encoder-decoder for statistical machine translation, 2014.

\bibitem[Cimpoi et~al.(2014)Cimpoi, Maji, Kokkinos, Mohamed, and Vedaldi]{DTD}
Mircea Cimpoi, Subhransu Maji, Iasonas Kokkinos, Sammy Mohamed, and Andrea Vedaldi.
\newblock Describing textures in the wild.
\newblock In \emph{Proceedings of the IEEE Conference on Computer Vision and Pattern Recognition (CVPR)}, 2014.

\bibitem[Corporation(2025)]{TE}
NVIDIA Corporation.
\newblock {Transformer Engine}, 2025.
\newblock Library for accelerating Transformer models on NVIDIA GPUs (FP8, FP16 mixed precision).

\bibitem[Dai et~al.(2019)Dai, Cai, Zhang, Xia, and Zhang]{ConvAttn-SAN}
Tao Dai, Jianrui Cai, Yongbing Zhang, Shu-Tao Xia, and Lei Zhang.
\newblock Second-order attention network for single image super-resolution.
\newblock In \emph{Proceedings of the IEEE/CVF Conference on Computer Vision and Pattern Recognition (CVPR)}, 2019.

\bibitem[Dong et~al.(2014)Dong, Loy, He, and Tang]{SRCNN}
Chao Dong, Chen~Change Loy, Kaiming He, and Xiaoou Tang.
\newblock Learning a deep convolutional network for image super-resolution.
\newblock In \emph{Computer Vision--ECCV 2014: 13th European Conference, Zurich, Switzerland, September 6-12, 2014, Proceedings, Part IV 13}, pages 184--199. Springer, 2014.

\bibitem[Dong et~al.(2016)Dong, Loy, and Tang]{Conv-FSRCNN}
Chao Dong, Chen~Change Loy, and Xiaoou Tang.
\newblock Accelerating the super-resolution convolutional neural network.
\newblock In \emph{Computer Vision -- ECCV 2016}, pages 391--407, Cham, 2016. Springer International Publishing.

\bibitem[Fuoli et~al.(2021)Fuoli, Van~Gool, and Timofte]{fuoli2021fourier}
Dario Fuoli, Luc Van~Gool, and Radu Timofte.
\newblock Fourier space losses for efficient perceptual image super-resolution.
\newblock In \emph{Proceedings of the IEEE/CVF International Conference on Computer Vision}, pages 2360--2369, 2021.

\bibitem[Guo et~al.(2017)Guo, Seyed~Mousavi, Huu~Vu, and Monga]{DWSR}
Tiantong Guo, Hojjat Seyed~Mousavi, Tiep Huu~Vu, and Vishal Monga.
\newblock Deep wavelet prediction for image super-resolution.
\newblock In \emph{Proceedings of the IEEE conference on computer vision and pattern recognition workshops}, pages 104--113, 2017.

\bibitem[Guo et~al.(2019)Guo, Mousavi, and Monga]{ORDSR}
Tiantong Guo, Hojjat~Seyed Mousavi, and Vishal Monga.
\newblock Adaptive transform domain image super-resolution via orthogonally regularized deep networks.
\newblock \emph{IEEE transactions on image processing}, 28\penalty0 (9):\penalty0 4685--4700, 2019.

\bibitem[He et~al.(2016)He, Zhang, Ren, and Sun]{ResNet}
Kaiming He, Xiangyu Zhang, Shaoqing Ren, and Jian Sun.
\newblock Deep residual learning for image recognition.
\newblock In \emph{Proceedings of the IEEE Conference on Computer Vision and Pattern Recognition (CVPR)}, 2016.

\bibitem[Hsu et~al.(2024)Hsu, Lee, and Chou]{DRCT}
Chih-Chung Hsu, Chia-Ming Lee, and Yi-Shiuan Chou.
\newblock Drct: Saving image super-resolution away from information bottleneck.
\newblock In \emph{Proceedings of the IEEE/CVF Conference on Computer Vision and Pattern Recognition}, pages 6133--6142, 2024.

\bibitem[Hu et~al.(2019)Hu, Mu, Zhang, Wang, Tan, and Sun]{MetaSR}
Xuecai Hu, Haoyuan Mu, Xiangyu Zhang, Zilei Wang, Tieniu Tan, and Jian Sun.
\newblock Meta-sr: A magnification-arbitrary network for super-resolution.
\newblock In \emph{Proceedings of the IEEE/CVF Conference on Computer Vision and Pattern Recognition (CVPR)}, 2019.

\bibitem[Huang et~al.(2017)Huang, He, Sun, and Tan]{WaveletSRNet}
Huaibo Huang, Ran He, Zhenan Sun, and Tieniu Tan.
\newblock Wavelet-srnet: A wavelet-based cnn for multi-scale face super resolution.
\newblock In \emph{Proceedings of the IEEE international conference on computer vision}, pages 1689--1697, 2017.

\bibitem[Huang et~al.(2015)Huang, Singh, and Ahuja]{Urban100}
Jia-Bin Huang, Abhishek Singh, and Narendra Ahuja.
\newblock Single image super-resolution from transformed self-exemplars.
\newblock In \emph{Proceedings of the IEEE Conference on Computer Vision and Pattern Recognition (CVPR)}, 2015.

\bibitem[Jeevan et~al.(2024)Jeevan, Srinidhi, Prathiba, and Sethi]{WaveMixSR}
Pranav Jeevan, Akella Srinidhi, Pasunuri Prathiba, and Amit Sethi.
\newblock Wavemixsr: Resource-efficient neural network for image super-resolution.
\newblock In \emph{Proceedings of the IEEE/CVF winter conference on applications of computer vision}, pages 5884--5892, 2024.

\bibitem[Kim et~al.(2016{\natexlab{a}})Kim, Lee, and Lee]{Conv-VDSR}
Jiwon Kim, Jung~Kwon Lee, and Kyoung~Mu Lee.
\newblock Accurate image super-resolution using very deep convolutional networks.
\newblock In \emph{Proceedings of the IEEE Conference on Computer Vision and Pattern Recognition (CVPR)}, 2016{\natexlab{a}}.

\bibitem[Kim et~al.(2016{\natexlab{b}})Kim, Lee, and Lee]{RecursiveSR-DRCN}
Jiwon Kim, Jung~Kwon Lee, and Kyoung~Mu Lee.
\newblock Deeply-recursive convolutional network for image super-resolution.
\newblock In \emph{Proceedings of the IEEE Conference on Computer Vision and Pattern Recognition (CVPR)}, 2016{\natexlab{b}}.

\bibitem[Lai et~al.(2017)Lai, Huang, Ahuja, and Yang]{LapSRN}
Wei-Sheng Lai, Jia-Bin Huang, Narendra Ahuja, and Ming-Hsuan Yang.
\newblock Deep laplacian pyramid networks for fast and accurate super-resolution.
\newblock In \emph{Proceedings of the IEEE conference on computer vision and pattern recognition}, pages 624--632, 2017.

\bibitem[Ledig et~al.(2017)Ledig, Theis, Husz{\'a}r, Caballero, Cunningham, Acosta, Aitken, Tejani, Totz, Wang, et~al.]{SRGAN}
Christian Ledig, Lucas Theis, Ferenc Husz{\'a}r, Jose Caballero, Andrew Cunningham, Alejandro Acosta, Andrew Aitken, Alykhan Tejani, Johannes Totz, Zehan Wang, et~al.
\newblock Photo-realistic single image super-resolution using a generative adversarial network.
\newblock In \emph{Proceedings of the IEEE conference on computer vision and pattern recognition}, pages 4681--4690, 2017.

\bibitem[Lee and Jin(2022)]{LTE}
Jaewon Lee and Kyong~Hwan Jin.
\newblock Local texture estimator for implicit representation function.
\newblock In \emph{Proceedings of the IEEE/CVF Conference on Computer Vision and Pattern Recognition (CVPR)}, pages 1929--1938, 2022.

\bibitem[Li et~al.(2022)Li, Lu, Qian, Lu, Zhang, and Jia]{EDT}
Wenbo Li, Xin Lu, Shengju Qian, Jiangbo Lu, Xiangyu Zhang, and Jiaya Jia.
\newblock On efficient transformer-based image pre-training for low-level vision, 2022.

\bibitem[Liang et~al.(2021)Liang, Cao, Sun, Zhang, Van~Gool, and Timofte]{SwinIR}
Jingyun Liang, Jiezhang Cao, Guolei Sun, Kai Zhang, Luc Van~Gool, and Radu Timofte.
\newblock Swinir: Image restoration using swin transformer.
\newblock In \emph{Proceedings of the IEEE/CVF International Conference on Computer Vision (ICCV) Workshops}, pages 1833--1844, 2021.

\bibitem[Lim et~al.(2017)Lim, Son, Kim, Nah, and Mu~Lee]{Conv-EDSR}
Bee Lim, Sanghyun Son, Heewon Kim, Seungjun Nah, and Kyoung Mu~Lee.
\newblock Enhanced deep residual networks for single image super-resolution.
\newblock In \emph{Proceedings of the IEEE Conference on Computer Vision and Pattern Recognition (CVPR) Workshops}, 2017.

\bibitem[Liu et~al.(2018{\natexlab{a}})Liu, Wen, Fan, Loy, and Huang]{RecursiveSR-nonlocal}
Ding Liu, Bihan Wen, Yuchen Fan, Chen~Change Loy, and Thomas~S Huang.
\newblock Non-local recurrent network for image restoration.
\newblock In \emph{Advances in Neural Information Processing Systems}. Curran Associates, Inc., 2018{\natexlab{a}}.

\bibitem[Liu et~al.(2020)Liu, Zhang, Tang, Tang, and Wu]{Conv-RFANet}
Jie Liu, Wenjie Zhang, Yuting Tang, Jie Tang, and Gangshan Wu.
\newblock Residual feature aggregation network for image super-resolution.
\newblock In \emph{Proceedings of the IEEE/CVF Conference on Computer Vision and Pattern Recognition (CVPR)}, 2020.

\bibitem[Liu et~al.(2018{\natexlab{b}})Liu, Zhang, Zhang, Lin, and Zuo]{MWCNN}
Pengju Liu, Hongzhi Zhang, Kai Zhang, Liang Lin, and Wangmeng Zuo.
\newblock Multi-level wavelet-cnn for image restoration.
\newblock In \emph{Proceedings of the IEEE conference on computer vision and pattern recognition workshops}, pages 773--782, 2018{\natexlab{b}}.

\bibitem[Liu et~al.(2021)Liu, Lin, Cao, Hu, Wei, Zhang, Lin, and Guo]{SwinTransformer}
Ze Liu, Yutong Lin, Yue Cao, Han Hu, Yixuan Wei, Zheng Zhang, Stephen Lin, and Baining Guo.
\newblock Swin transformer: Hierarchical vision transformer using shifted windows.
\newblock In \emph{Proceedings of the IEEE/CVF International Conference on Computer Vision (ICCV)}, pages 10012--10022, 2021.

\bibitem[Martin et~al.(2001)Martin, Fowlkes, Tal, and Malik]{BSDS100}
D. Martin, C. Fowlkes, D. Tal, and J. Malik.
\newblock A database of human segmented natural images and its application to evaluating segmentation algorithms and measuring ecological statistics.
\newblock In \emph{Proceedings Eighth IEEE International Conference on Computer Vision. ICCV 2001}, pages 416--423 vol.2, 2001.

\bibitem[Matsui et~al.(2017)Matsui, Ito, Aramaki, Fujimoto, Ogawa, Yamasaki, and Aizawa]{Manga109}
Yusuke Matsui, Kota Ito, Yuji Aramaki, Azuma Fujimoto, Toru Ogawa, Toshihiko Yamasaki, and Kiyoharu Aizawa.
\newblock Sketch-based manga retrieval using manga109 dataset.
\newblock \emph{Multimedia Tools and Applications}, 76\penalty0 (20):\penalty0 21811--21838, 2017.

\bibitem[Mei et~al.(2021)Mei, Fan, and Zhou]{ConvAttn-NLSN}
Yiqun Mei, Yuchen Fan, and Yuqian Zhou.
\newblock Image super-resolution with non-local sparse attention.
\newblock In \emph{Proceedings of the IEEE/CVF Conference on Computer Vision and Pattern Recognition (CVPR)}, pages 3517--3526, 2021.

\bibitem[Mescheder et~al.(2019)Mescheder, Oechsle, Niemeyer, Nowozin, and Geiger]{INR_1}
Lars Mescheder, Michael Oechsle, Michael Niemeyer, Sebastian Nowozin, and Andreas Geiger.
\newblock Occupancy networks: Learning 3d reconstruction in function space.
\newblock In \emph{Proceedings of the IEEE/CVF conference on computer vision and pattern recognition}, pages 4460--4470, 2019.

\bibitem[Niu et~al.(2020)Niu, Wen, Ren, Zhang, Yang, Wang, Zhang, Cao, and Shen]{ConvAttn-HAN}
Ben Niu, Weilei Wen, Wenqi Ren, Xiangde Zhang, Lianping Yang, Shuzhen Wang, Kaihao Zhang, Xiaochun Cao, and Haifeng Shen.
\newblock Single image super-resolution via a holistic attention network.
\newblock In \emph{Computer Vision -- ECCV 2020}, pages 191--207, Cham, 2020. Springer International Publishing.

\bibitem[Odena et~al.(2016)Odena, Dumoulin, and Olah]{NNConv}
Augustus Odena, Vincent Dumoulin, and Chris Olah.
\newblock Deconvolution and checkerboard artifacts.
\newblock \emph{Distill}, 2016.

\bibitem[Park et~al.(2019)Park, Florence, Straub, Newcombe, and Lovegrove]{INR_0}
Jeong~Joon Park, Peter Florence, Julian Straub, Richard Newcombe, and Steven Lovegrove.
\newblock Deepsdf: Learning continuous signed distance functions for shape representation.
\newblock In \emph{Proceedings of the IEEE/CVF conference on computer vision and pattern recognition}, pages 165--174, 2019.

\bibitem[Park et~al.(2003)Park, Park, and Kang]{Park2003}
Sung~Cheol Park, Min~Kyu Park, and Moon~Gi Kang.
\newblock Super-resolution image reconstruction: a technical overview.
\newblock \emph{IEEE Signal Processing Magazine}, 20\penalty0 (3):\penalty0 21--36, 2003.

\bibitem[Rosenthal and Henderson(2003)]{FRC_1}
Peter~B Rosenthal and Richard Henderson.
\newblock Optimal determination of particle orientation, absolute hand, and contrast loss in single-particle electron cryomicroscopy.
\newblock \emph{Journal of molecular biology}, 333\penalty0 (4):\penalty0 721--745, 2003.

\bibitem[Saxton(1982)]{FRC_0}
WO Saxton.
\newblock The correlation averaging of a regularly arranged bacterial cell envelope protein.
\newblock \emph{Journal of microscopy}, 127\penalty0 (2):\penalty0 127--138, 1982.

\bibitem[Shi et~al.(2016)Shi, Caballero, Huszar, Totz, Aitken, Bishop, Rueckert, and Wang]{ESPCN}
Wenzhe Shi, Jose Caballero, Ferenc Huszar, Johannes Totz, Andrew~P. Aitken, Rob Bishop, Daniel Rueckert, and Zehan Wang.
\newblock Real-time single image and video super-resolution using an efficient sub-pixel convolutional neural network.
\newblock In \emph{Proceedings of the IEEE Conference on Computer Vision and Pattern Recognition (CVPR)}, 2016.

\bibitem[Timofte et~al.(2017)Timofte, Agustsson, Van~Gool, Yang, and Zhang]{Flick2K}
Radu Timofte, Eirikur Agustsson, Luc Van~Gool, Ming-Hsuan Yang, and Lei Zhang.
\newblock Ntire 2017 challenge on single image super-resolution: Methods and results.
\newblock In \emph{Proceedings of the IEEE conference on computer vision and pattern recognition workshops}, pages 114--125, 2017.

\bibitem[Tong et~al.(2017)Tong, Li, Liu, and Gao]{Conv-SRDenseNet}
Tong Tong, Gen Li, Xiejie Liu, and Qinquan Gao.
\newblock Image super-resolution using dense skip connections.
\newblock In \emph{Proceedings of the IEEE International Conference on Computer Vision (ICCV)}, 2017.

\bibitem[Van~Heel and Schatz(2005)]{FRC_2}
Marin Van~Heel and Michael Schatz.
\newblock Fourier shell correlation threshold criteria.
\newblock \emph{Journal of structural biology}, 151\penalty0 (3):\penalty0 250--262, 2005.

\bibitem[Wang et~al.(2021)Wang, Wang, Lin, Yang, An, and Guo]{ArbRCAN}
Longguang Wang, Yingqian Wang, Zaiping Lin, Jungang Yang, Wei An, and Yulan Guo.
\newblock Learning a single network for scale-arbitrary super-resolution.
\newblock In \emph{Proceedings of the IEEE/CVF International Conference on Computer Vision (ICCV)}, pages 4801--4810, 2021.

\bibitem[Wang et~al.(2018)Wang, Yu, Wu, Gu, Liu, Dong, Qiao, and Change~Loy]{ESRGAN}
Xintao Wang, Ke Yu, Shixiang Wu, Jinjin Gu, Yihao Liu, Chao Dong, Yu Qiao, and Chen Change~Loy.
\newblock Esrgan: Enhanced super-resolution generative adversarial networks.
\newblock In \emph{Proceedings of the European conference on computer vision (ECCV) workshops}, pages 0--0, 2018.

\bibitem[Wang et~al.(2022)Wang, Cun, Bao, Zhou, Liu, and Li]{Uformer}
Zhendong Wang, Xiaodong Cun, Jianmin Bao, Wengang Zhou, Jianzhuang Liu, and Houqiang Li.
\newblock Uformer: A general u-shaped transformer for image restoration.
\newblock In \emph{Proceedings of the IEEE/CVF Conference on Computer Vision and Pattern Recognition (CVPR)}, pages 17683--17693, 2022.

\bibitem[Yang et~al.(2018)Yang, Mei, Zhang, Xu, Yin, Zhang, and Wei]{DRFN}
Xin Yang, Haiyang Mei, Jiqing Zhang, Ke Xu, Baocai Yin, Qiang Zhang, and Xiaopeng Wei.
\newblock Drfn: Deep recurrent fusion network for single-image super-resolution with large factors.
\newblock \emph{IEEE Transactions on Multimedia}, 21\penalty0 (2):\penalty0 328--337, 2018.

\bibitem[Zeiler et~al.(2010)Zeiler, Krishnan, Taylor, and Fergus]{Deconv}
Matthew~D. Zeiler, Dilip Krishnan, Graham~W. Taylor, and Rob Fergus.
\newblock Deconvolutional networks.
\newblock In \emph{2010 IEEE Computer Society Conference on Computer Vision and Pattern Recognition}, pages 2528--2535, 2010.

\bibitem[Zeyde et~al.(2012)Zeyde, Elad, and Protter]{Set14}
Roman Zeyde, Michael Elad, and Matan Protter.
\newblock On single image scale-up using sparse-representations.
\newblock In \emph{Curves and Surfaces}, pages 711--730, Berlin, Heidelberg, 2012. Springer Berlin Heidelberg.

\bibitem[Zhang et~al.(2018{\natexlab{a}})Zhang, Li, Li, Wang, Zhong, and Fu]{ConvAttn-RCAN}
Yulun Zhang, Kunpeng Li, Kai Li, Lichen Wang, Bineng Zhong, and Yun Fu.
\newblock Image super-resolution using very deep residual channel attention networks.
\newblock In \emph{Proceedings of the European Conference on Computer Vision (ECCV)}, 2018{\natexlab{a}}.

\bibitem[Zhang et~al.(2018{\natexlab{b}})Zhang, Tian, Kong, Zhong, and Fu]{Conv-RDN}
Yulun Zhang, Yapeng Tian, Yu Kong, Bineng Zhong, and Yun Fu.
\newblock Residual dense network for image super-resolution.
\newblock In \emph{Proceedings of the IEEE Conference on Computer Vision and Pattern Recognition (CVPR)}, 2018{\natexlab{b}}.

\bibitem[Zhang et~al.(2019)Zhang, Li, Li, Zhong, and Fu]{ConvAttn-RNAN}
Yulun Zhang, Kunpeng Li, Kai Li, Bineng Zhong, and Yun Fu.
\newblock Residual non-local attention networks for image restoration, 2019.

\end{thebibliography}
